\documentclass[10pt,journal,compsoc]{IEEEtran}
\usepackage{multirow}
\usepackage[export]{adjustbox}
\usepackage{mathtools}
\usepackage{xcolor}
\usepackage{amsmath}
\usepackage{amssymb}
\usepackage{hyperref}
\usepackage[nolist]{acronym}
\usepackage{subcaption}
\usepackage{units}
\usepackage{pifont}
\usepackage{xstring}
\usepackage[normalem]{ulem}

\newcommand\sergio[1]{}
\newcommand\oswald[1]{}
\newcommand\swathi[1]{}
\newcommand\olrev[2]{}
\newcommand\rev[2]{}
\newcommand\newtext[1]{#1}

\newcommand{\sota}{state-of-the-art }

\usepackage[normalem]{ulem}

\newcommand\revnew[2]{#2}

\newcommand\revnewv[2]{#2}

\newcommand\revnewvt[2]{#2}

\definecolor{gsf_green}{RGB}{112,173,71}

\ifCLASSOPTIONcompsoc
\usepackage[nocompress]{cite}
\else
\usepackage{cite}
\fi
\ifCLASSINFOpdf
\else
\fi

\newacro{lstm}[LSTM]{Long Short-Term Memory}
\newacro{lsta}[LSTA]{Long Short-Term Attention}
\newacro{clstm}[ConvLSTM]{Convolutional Long Short-Term Memory}
\newacro{rnn}[RNN]{Recurrent Neural Network}
\newacro{cnn}[CNN]{Convolutional Neural Network}
\newacro{tsn}[TSN]{Temporal Segment Network}
\newacro{trn}[TRN]{Temporal Relation Network}
\newacro{tdn}[TDN]{Temporal Difference Network}
\newacro{mfn}[MFNet]{Motion Feature Network}
\newacro{fc}[FC]{Fully Connected}
\newacro{bn}[BN]{Batch Normalization}
\newacro{gru}[GRU]{Gated Recurrent Unit}
\newacro{sgd}[SGD]{Stochastic Gradient Descent}
\newacro{tsm}[TSM]{Temporal Shift Module}
\newacro{flop}[FLOP]{Floating Point Operation}
\newacro{lrcn}[LRCN]{Long-term Recurrent Convolutional Network}
\newacro{gsm}[GSM]{Gate-Shift Module}
\newacro{gsf}[GSF]{Gate-Shift-Fuse}
\newacro{gsn}[GSN]{Gate-Shift-Fuse Network}
\newacro{gst}[GST]{Grouped Spatial-temporal Aggregation}

\newcommand\ie{\emph{i.e}.}

\newcommand\etal{\emph{et al}. }
\newcommand\etc{\emph{etc}.}

\graphicspath{{figs/}}

\begin{document}
	
	%
	\title{Gate-Shift-Fuse for Video Action Recognition}
	%
	%
	%
	%
	
	\author{Swathikiran~Sudhakaran, 
		Sergio~Escalera, 
		and~Oswald~Lanz
		\IEEEcompsocitemizethanks{
			\IEEEcompsocthanksitem S. Sudhakaran was with FBK, Trento, Italy with this work and is now with Samsung AI Center, Cambridge, UK. E-mail: swathikirans@gmail.com
			\IEEEcompsocthanksitem S. Escalera is with Universitat de Barcelona and Computer Vision Center, Spain. E-mail: sescalera@ub.edu
			\IEEEcompsocthanksitem O. Lanz was with FBK, Trento, Italy with this work and is now with Free University of Bozen-Bolzano, Italy. E-mail: oswald.lanz@unibz.it}
		\thanks{Manuscript submitted October 28, 2021; revised August 9, 2022, revised December 20, 2022; revised April 7, 2023}}
	
	%
	%

	\markboth{Accepted at IEEE TPAMI}%
	{}
	%



	\IEEEtitleabstractindextext{%
		\begin{abstract}
			\newtext{Convolutional Neural Networks are the de facto models for image recognition. However 3D CNNs, the straight forward extension of 2D CNNs for video recognition, have not achieved the same success on standard action recognition benchmarks. One of the main reasons for this reduced performance of 3D CNNs is the increased computational complexity requiring large scale annotated datasets to train them in scale. 3D kernel factorization approaches have been proposed to reduce the complexity of 3D CNNs. Existing kernel factorization approaches follow hand-designed and hard-wired techniques. In this paper we propose Gate-Shift-Fuse (GSF), a novel spatio-temporal feature extraction module which controls interactions in spatio-temporal decomposition and learns to adaptively route features through time and combine them in a data dependent manner. GSF leverages grouped spatial gating to decompose input tensor and channel weighting to fuse the decomposed tensors. GSF can be inserted into existing 2D CNNs to convert them into an efficient and high performing spatio-temporal feature extractor, with negligible parameter and compute overhead. We perform an extensive analysis of GSF using two popular 2D CNN families and achieve \sota or competitive performance on five standard action recognition benchmarks.}
		\end{abstract}
		
		}

		\maketitle

		\IEEEdisplaynontitleabstractindextext

		%
		\IEEEpeerreviewmaketitle
		
		\IEEEraisesectionheading{\section{Introduction}\label{sec:intro}}
		
\newtext{Understanding human actions in videos is one of the challenging and fundamental problems in computer vision with a wide range of applications from video surveillance to robotics and human-computer interaction. Recently, video action recognition research has witnessed great progress owing to the adoption of Deep Neural Networks. However, such Deep Neural Models are yet to achieve the success of their image recognition counterparts. A key challenge lies in addressing the question of how to better encode the space-time features.} 

\newtext{Inspired from the improved performance of 2D~\acp{cnn} on image recognition tasks, \cite{karpathy2014large} explores several approaches to extend 2D~\acp{cnn} for action recognition by considering videos as a set of frames. This includes using a single frame from the video to stacking multiple frames and applying it to a 2D~\ac{cnn}. However it is found that both of these configurations perform similarly, indicating that the learned video representation is not capable of capturing motion information in the video. Two-Stream networks~\cite{twoStream},~\cite{feichtenhofer2016convolutional} address this problem by stacking optical flow images to capture short-term motion features and incorporate them with the appearance features obtained from a single frame. Such approaches require costly optical flow computations and ignore valuable appearance features from the video. TSN~\cite{tsn} and ActionVLAD~\cite{vlad} propose to extract features from multiple frames instead, followed by a late temporal aggregation via mean pooling for improved appearance feature extraction. Such approaches still relied on optical flow images for capturing motion information. The works of~\cite{lrcn},~\cite{vlstm},~\cite{lsta} formulate late temporal aggregation as a sequence learning problem by encoding the frame-level features using variants of \acp{rnn}. All these approaches perform temporal modelling in a shallow manner.}

\begin{figure}
\includegraphics[width=.95\columnwidth]{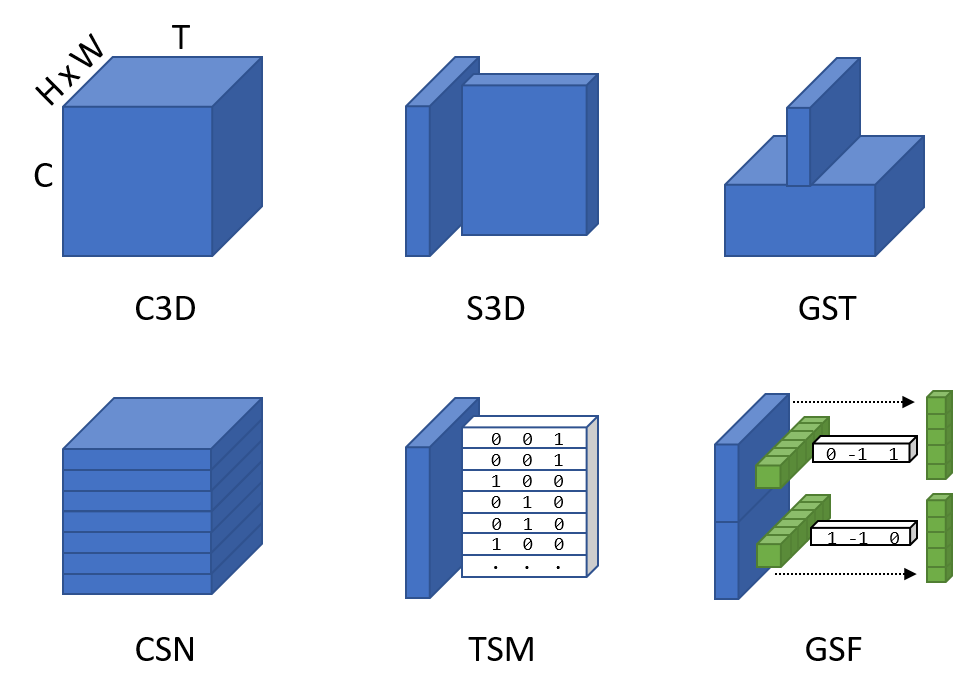}
\vspace{-0.3cm}
\caption{3D kernel factorization for spatio-temporal learning in video. \revnewv{Existing}{Majority of previous} approaches decompose into channel-wise (CSN), spatial followed by temporal (S3D, TSM), or grouped spatial and spatio-temporal (GST). In \revnewv{all} these, spatial, temporal, and channel-wise interaction is hard-wired. Our {\bf Gate-Shift-Fuse (GSF)} leverages group spatial gating and fusion (blocks in green) to control interactions in spatial-temporal decomposition. GSF is lightweight and a building block of high performing video feature extractors.
}
\label{fig:kernel}
\end{figure}

Fine-grained recognition can benefit from deeper temporal modeling. Full-3D CNNs (C3D~\cite{tran2015learning,hara2018can}) process the video in space-time by
expanding the kernels of a 2D ConvNet along the temporal dimension. Deep C3Ds are designed to learn powerful representations in the joint spatio-temporal feature space with more parameters (3D kernels) and computations (kernels slide over 3 densely sampled dimensions). In practice however, they may under-perform due to the lack of sufficiently large datasets for training them at scale. To cope with these issues arising from the curse of dimension one can narrow down network capacity by design. Fig.~\ref{fig:kernel} shows several C3D kernel decomposition approaches proposed for spatio-temporal feature learning in video. A most intuitive approach is to factorize 3D spatio-temporal kernels into 2D spatial plus 1D temporal, 
resulting in a structural decomposition that disentangles spatial from temporal interactions (P3D~\cite{p3d}, R(2+1)D~\cite{r2plus1d}, S3D~\cite{s3d}). An alternative design is separating channel interactions and spatio-temporal interactions via group convolution (CSN~\cite{csn}), or modeling both spatial and spatio-temporal interactions in parallel with 2D and 3D convolution on separated channel groups (GST~\cite{gst}). Temporal convolution can be constrained to hard-coded time-shifts that move some of the channels forward in time or backward (TSM~\cite{tsm}). \revnewv{All these existing}{These} approaches learn structured kernels with a hard-wired connectivity and propagation pattern across the network. There is no data dependent decision taken at any point in the network to route features selectively through different branches, for example, group-and-shuffle patterns are fixed by design and learning how to shuffle is combinatorial complexity. \revnewv{}{Recently,~\cite{rubiksnet} introduced RubiksNet, a special case of TSM, that replaces spatial convolution and temporal shifting with learnable shifting operations to achieve a flexible spatio-temporal decomposition.}

In this paper we introduce spatial gating \newtext{and channel fusion} in spatial-temporal decomposition of 3D kernels. We implement this concept with \ac{gsf} as shown in Fig.~\ref{fig:kernel}. \ac{gsf} is lightweight and turns a 2D-CNN into a highly efficient spatio-temporal feature extractor. \ac{gsf} first applies 2D convolution, then decomposes the output tensor using a learnable spatial gating into two tensors: a gated version of it, and its residual. The gated tensor goes through a 1D temporal convolution while its residual is skip-connected to its output. We implement spatial gating as group spatio-temporal convolution with single output plane per group. We use hard-coded time-shift of channel groups instead of learnable temporal convolution. \newtext{Channel fusion is implemented by generating channel weights via a light weight 2D convolution. The two feature groups, gate-shifted and residual, are then fused by weighted averaging across the channels.} With \ac{gsf} plugged in, a 2D-CNN learns to adaptively route features through time and combine them, at almost no additional parameters and computational overhead. For example, when \ac{gsf} is plugged into TSN~\cite{tsn}, an absolute gain of $+32$ percentage points in accuracy is obtained on Something Something-V1 dataset with just $0.48\%$ additional parameters and $0.58\%$ additional \acp{flop}. 

The contributions of this paper can be summarized as follows: 
\begin{itemize}
	\item We propose a novel spatio-temporal feature extraction module that can be plugged into existing 2D \ac{cnn} architectures with negligible overhead in terms of computations and memory;
	\item We perform an extensive ablation analysis of the proposed module to study its effectiveness in video action recognition;
	\item \newtext{We perform an extensive analysis to study how to effectively integrate the proposed module with two popular 2D~\ac{cnn} architecture families and} achieve \sota or competitive results on five public benchmarks with less parameters and \acp{flop} compared to existing approaches.
\end{itemize}

\revnew{}{This paper extends our earlier work~\cite{gsm} published in CVPR 2020 in many aspects. In particular, we introduce a weighted fusion of the two feature groups, gate-shifted and its residual, instead of the naive average fusion presented in~\cite{gsm}. Secondly, we present an extensive analysis to identify the best design choice for applying our light weight module to ResNet based backbone architectures. Thirdly, we evaluate the performance of our module on two additional datasets: Kinetics 400 and Something Something-v2 and replace two datasets (EPIC-Kitchens-55 and Diving48) used in~\cite{gsm} with their updated versions. We also extend the related works section by reviewing more recent works on video action recognition and the experimental section by performing an extensive comparison of the performance of our work with more recently developed approaches.}

The rest of the paper is organized as follows. Sec.~\ref{sec:related_works} reviews recent works on action recognition. Sec.~\ref{sec:gsf} presents the technical details of \acf{gsf}. Experimental results are reported in Sec.~\ref{sec:experiments} and Sec.~\ref{sec:conclusion} concludes the paper.

		
		\section{Related Work}
		\label{sec:related_works}
		
\newtext{The development of large-scale video datasets such as Kinetics~\cite{kin400}~\cite{kin700}, Moments in Time~\cite{momentsintime}, Something Something~\cite{goyal2017something}, EPIC-Kitchens-100~\cite{epic100}, \etc, to mention a few, inspired video action recognition research significantly. Such datasets with varying properties, such as requirement of spatial reasoning~\cite{kin400}, temporal reasoning~\cite{goyal2017something}, spatio-temporal reasoning~\cite{epic100}, resulted in the development of a plethora of action recognition approaches. The development of recent large-scale datasets such as HowTo100M~\cite{howto100m}, HVU~\cite{hvu}, Ego4D~\cite{ego4d}, \etc, are expected to further video understanding research beyond simple tasks such as action recognition.}

\subsection{Action Recognition}
\vspace*{3pt}
\noindent{\bf Fusing appearance and flow.} 
A popular extension of 2D CNNs to handle video is the Two-Stream architecture by Simonyan and Zisserman~\cite{twoStream}. Their method consists of two separated CNNs (streams) that are trained to extract features from a sampled RGB video frame paired with the surrounding stack of optical flow images, followed by a late fusion of the prediction scores of both streams. The image stream encodes the appearance information while the optical flow stream encodes the motion information, that are often found to complement each other for action recognition. 
Several works followed this approach to find a suitable fusion of the streams at various depths~\cite{feichtenhofer2016convolutional} and to explore the use of residual connections~\cite{feichtenhofer2016spatiotemporal} \newtext{or multiplicative gating functions~\cite{feichtenhofer2017spatiotemporal}} between them. These approaches rely on optical flow images for motion information, and a single RGB frame for appearance information, which is limiting when reasoning about the temporal context is required for video understanding.

\vspace*{3pt}
\noindent{\bf Video as a set or sequence of frames.} Later, other approaches were developed using multiple RGB frames for video classification. These approaches sparsely sample multiple frames from the video, which are applied to a 2D \ac{cnn} followed by a late integration of frame-level features using average pooling~\cite{tsn}, multilayer perceptrons~\cite{trn}, recurrent aggregation~\cite{lrcn, vlstm}, or attention~\cite{pool,lsta}. To boost performance, most of these approaches also combine video frame sequence with externally computed optical flow. This shows to be helpful, but computationally intensive.

\vspace*{3pt}
\noindent{\bf Modeling short-term temporal dependencies.} 
Other research has investigated the middle ground between late aggregation (of frame features) and early temporal processing (to get optical flow), by modeling short-term dependencies.
This includes differencing of intermediate features~\cite{tdn} and combining Sobel filtering with feature differencing~\cite{off}. Other works~\cite{tvnet, piergiovanni2019representation} develop a differentiable network that performs TV-L1~\cite{tv-l1}, a popular optical flow extraction technique. The work of \cite{mfnet} instead uses a set of fixed filters for extracting motion features, thereby greatly reducing the number of parameters. DMC-Nets~\cite{dmc} leverage motion vectors in the compressed video to synthesize discriminative motion cues for two-stream action recognition at low computational cost compared to raw flow extraction.

\vspace*{3pt}
\noindent{\bf Video as a space-time volume.} Unconstrained modeling and learning of action features is possible when considering video in space-time.
Since video can be seen as a temporally dense sampled sequence of images, expanding 2D convolution operation in 2D-CNNs to 3D convolution is a most intuitive approach to spatio-temporal feature learning~\cite{tran2015learning, hara2018can, carreira2017quo}. The major drawback of 3D \acp{cnn} is the huge number of parameters involved. This results in increased computations and the requirement of large scale datasets for pre-training. Carreira and Zisserman~\cite{carreira2017quo} addressed this limitation by inflating video 3D kernels with the 2D weights of a \ac{cnn} trained for image recognition. Several other approaches focused on reducing the number of parameters by disentangling the spatial and temporal feature extraction operations. 
P3D~\cite{p3d} proposes three different choices for separating the spatial and temporal convolutions and develops a 3D-ResNet architecture whose residual units are a sequence of such three modules. R(2+1)D~\cite{r2plus1d} and S3D-G~\cite{s3d} also show that a 2D convolution followed by 1D convolution is enough to learn discriminative features for action recognition.
CoST~\cite{cost} performs 2D convolutions, with shared parameters, along the three orthogonal dimensions of a video sequence. \newtext{CT-Net~\cite{ctnet} proposes to tensorize the channel dimension of features as a multiplication of K sub-dimensions, followed by 3D separable convolutions along each of the K channel sub-dimensions. This allows the model to gradually increase the spatio-temporal receptive field with reduced computational complexity while maintaining interaction of features across both spatio-temporal and channel dimensions.} MultiFiber~\cite{multifiber} uses multiple lightweight networks, the fibers, and multiplexer modules that facilitate information flow using point-wise convolutions across the fibers. \newtext{CIDC~\cite{cidc} performs channel independent directional convolution by applying 2D convolution independently on individual feature channels while considering the temporal dimension as channels. Other works involve  Neural Architecture Search (NAS) for developing an optimum architecture~\cite{assemblenet},~\cite{assemblenet++},~\cite{attentionnas}. X3D~\cite{x3d} develops a progressive algorithm for expanding the spatio-temporal resolution of inputs, network depth, number of channels, \etc, of 3D~\ac{cnn} architectures. The final architecture developed with the search algorithm results in improved performance with less computational complexity compared to hand-designed 3D~\acp{cnn}.} 

\vspace*{3pt}
\noindent{\newtext{\bf Modeling long-term dependencies.}} \newtext{Majority of existing approaches rely on the network depth to combine the information present in frames that are further apart in time. \cite{lfb} develops a feature bank that stores features from the entire video which can then be integrated as contextual information to a video recognition model for understanding the present. Non-local~\cite{nonlocal} is a plug and play module that computes the features at a specific position in space-time as the weighted sum of features at all other positions, thereby capturing long-range dependencies. Timeception~\cite{timeception} introduces multi-scale separable temporal convolutions to reason about videos of longer duration. Eidectic~3D~LSTM~\cite{wang2018eidetic} integrates 3D convolution into \acsp{rnn} and introduces a gated controlled self-attention module to enable the memory to interact with its past values. The former enables short-term temporal modeling while the latter allows the model to encode long-term temporal information. VideoGraph~\cite{hussein2019videograph} constructs a unidirectional graph whose nodes represent the latent action concepts. A novel graph embedding layer is then developed to learn the relationship between these concepts. Zhang~\etal~\cite{v4d} introduces 4D convolution to perform long-term modeling hierarchically. Inspired by the performance improvement obtained by Transformers in Natural Language Processing, Vision Transformer (ViT)~\cite{vit} is developed for image recognition. TimeSformer~\cite{timesformer} extends ViT by developing different variants of self-attention schemes for spatio-temporal feature learning. However, even the efficient variant of TimeSformer suffers from increased computational complexity and larger parameter count.} 

\begin{figure*}[h]
	\centering\includegraphics[width=0.8\linewidth]{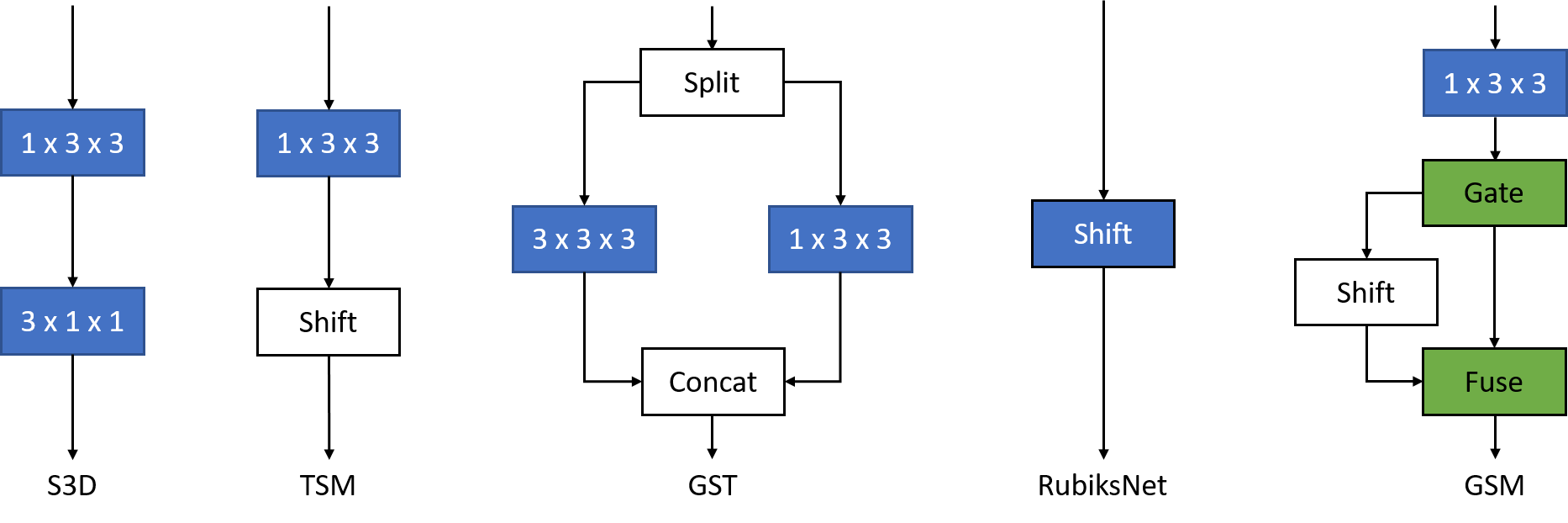}\vspace{-0.4cm}
	\caption{C3D~\cite{tran2015learning}  decomposition approaches in comparison to GSF schematics. \revnewvt{}{S3D~\cite{s3d} decomposes C3D into spatial and temporal convolutions while TSM~\cite{tsm} replaces the 1D temporal convolution with fixed channel-wise shifting. GST~\cite{gst} decomposes C3D into grouped spatial and spatio-temporal convolutions by splitting across the channel dimension.} \revnewvt{}{RubiksNet~\cite{rubiksnet} implements learnable spatio-temporal shifting to approximate expensive 3D convolution operation.}  GSF is inspired by GST and TSM but replaces the hard-wired channel split and concat aggregation with learnable \textbf{\textcolor{gsf_green}{spatial gating}} and \textbf{\textcolor{gsf_green}{fusion}} blocks. \revnewvt{}{Compared to RubiksNet that needs to be trained from scratch, GSF can be plugged into any existing 2D CNNs thereby leveraging ImageNet pre-trained weights.}}
	\label{fig:comp}
\end{figure*}

\vspace*{3pt}
\noindent{\newtext{\bf Multi-scale architectures.}} \newtext{Inspired by the effectiveness of Two-stream architectures using RGB and optical flow images,~\cite{karpathy2014large}~\cite{slowfast},~\cite{blvnet-tam} developed multi-stream architectures operating on two streams of RGB images. A multi-resolution \ac{cnn} architecture is developed in~\cite{karpathy2014large} where low resolution frames are processed by one stream and high resolution center crops are processed by the second stream. \cite{blvnet-tam} proposes to apply alternate frames of high resolution through a compact network in addition to a second stream processing low resolution images. This allows the network to encode features at multiple spatial scales and increase the number of processed frames while maintaining reasonable computational complexity. SlowFast~\cite{slowfast} encodes frames using a dual pathway network. One pathway encodes frames sampled at a slower frame rate while the second encodes frames sampled at a faster frame rate, enabling the network to capture information at multiple temporal scales. These approaches rely on multiple networks to encode information at different scales. To address this issue, TPN~\cite{tpn}  introduces a feature level pyramid network to encode features at multiple hierarchies thereby allowing the network to extract features at different temporal scales. Since the pyramid structure is constructed at feature level, TPN do not require multiple architectures, thereby reducing the number of parameters and computational complexity. MSTI-Net~\cite{wu2020multi} proposes to split the feature tensors across the channel dimension into four groups and apply convolution operation with different kernel size and aggregate them hierarchically. Varying the kernel size of the convolution operation enables the network to encode spatio-temporal features at multiple scales.}

\subsection{Efficient Video Understanding}

Recently, the focus of research is moving to the development of efficient (from a computational point of view) and effective (from a performance point of view) approaches for video understanding. \newtext{Such approaches perform efficient modeling by (1) reducing redundant computations or (2) developing novel efficient architectures.}

\vspace*{3pt}
\noindent{\newtext{\bf Reducing computations by selective processing.}} \newtext{\cite{coarse-fine} and \cite{liteeval} develop a conditional computing approach that decides whether to compute fine resolution features in a data dependent manner, to avoid additional compute overhead. AdaFrame~\cite{adaframe} trains a policy gradient network to determine which frame to process next, thereby skipping processing of non-informative frames in a video. SCSampler~\cite{scsampler} develops a light-weight model for identifying the most salient clip for recognizing the action present in a video. AdaFuse~\cite{adafuse} learns a decision policy to reduce temporal redundancy by removing feature channels that are uninformative and reuse the feature channels from past frames for efficient inference. A reinforcement policy based learning approach for selecting the optimum input frame resolution is developed in~\cite{arnet}. \cite{framedistillation} follows the approach of knowledge distillation to train a student network that processes a smaller number of frames using a teacher network trained with a larger number of frames. All these approaches are based on existing spatio-temporal feature extractors or architectures that require large memory requirements due to the model size. Even though these approaches reduce the computations in inference time, majority of them suffer from heavy computations during the training stage.}

\vspace*{3pt}
\noindent{\newtext{\bf Efficient architectures.}} \acp{cnn} provide different levels of feature abstractions at different layers of the hierarchy. It has been found that the bottom layer features are less useful for extracting discriminative motion cues~\cite{sun2015human, eco, s3d}. In \cite{sun2015human} it is proposed to apply 1D convolution layers on top of a 2D \ac{cnn} for video action recognition. The works of \cite{eco} and \cite{s3d} show that it is more effective to apply full 3D and separable 3D convolutions at the top layers of a 2D \ac{cnn} for extracting spatio-temporal features. These approaches resulted in performance improvement over full 3D architectures with less parameters and computations. Static features from individual frames represent scenes and objects and can also provide important cues in identifying the action. This is validated by the improved performance obtained with two-path structures that apply a parallel 2D convolution in addition to the 3D convolution~\cite{mict, gst}. MiCT~\cite{mict} is designed by adding 3D convolution branches in parallel to the 2D convolution branches of a BN-Inception-like \ac{cnn}. GST~\cite{gst} makes use of the idea of grouped convolutions for developing an efficient architecture for action recognition. They separate the features at a hierarchy across the channel dimension and separately perform 2D and 3D convolutions followed by a concatenation operation. In this way, the performance is increased while reducing the number of parameters. \newtext{MVFNet~\cite{mvfnet} extends the idea of GST by replacing the 3D convolution with three 1D convolutions applied to each of the two spatial dimensions and temporal dimension.} STM~\cite{stm} proposes two parallel blocks for extracting motion features and spatio-temporal features. Their network rely only on 2D and 1D convolutions and feature differencing for encoding motion and spatio-temporal features. TSM~\cite{tsm} proposes to shift the features across the channel dimension as a way to perform temporal interaction between the features from adjacent frames of a video. This parameter-less approach has resulted in similar performance to 3D \acp{cnn}. \newtext{RubiksNet~\cite{rubiksnet} replaces the convolution operation in 3D~\acp{cnn} with learnable spatio-temporal shifting. TEA~\cite{tea} develops a novel ResNet based architecture that weights channels with motion information while suppressing those without relevant information and that can encode long range temporal information in a hierarchical aggregation fashion. MSNet~\cite{motionsqueeze} develops a plug in module that can estimate the correspondences between adjacent frame level features and convert these into effective motion features that can be injected into the backbone \ac{cnn} for efficient spatio-temporal modeling.} 

In all previous approaches, spatial, temporal, and channel-wise  interaction is hard-wired. Here, we propose the \acf{gsf}, which control interactions in spatial-temporal decomposition and learns to adaptively route features through time and combine them \newtext{in a data dependent manner}, at almost no additional parameters and computational overhead.

		\section{Gate-Shift\newtext{-Fuse}}
		\label{sec:gsf}

\newtext{We present \acf{gsf}, a module capable of converting a 2D~\ac{cnn} into a high performing spatio-temporal feature extractor with minimal overhead.}

\subsection{\acl{gsf} Module}
\label{sec:gsf_module}

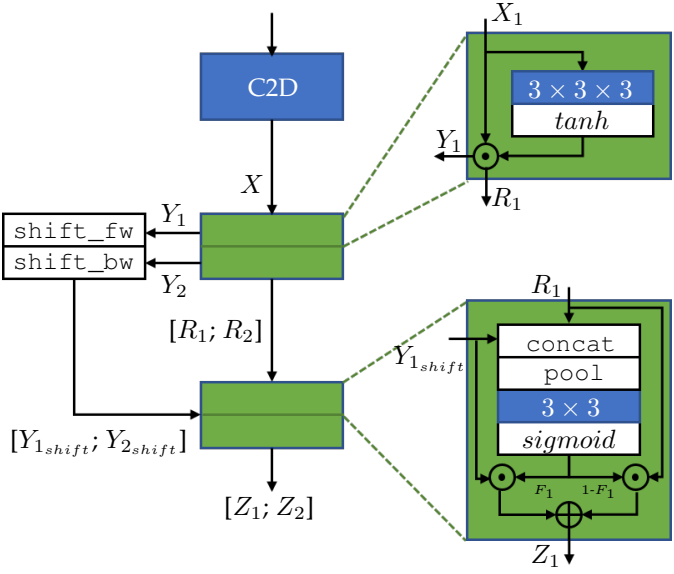
\begin{figure}
	\centering
	\includegraphics[width=\columnwidth]{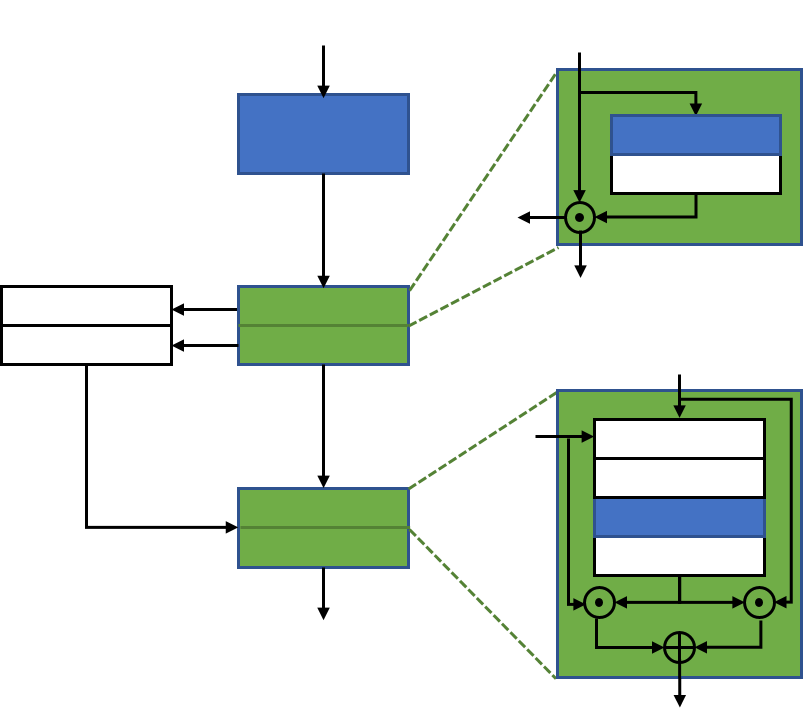}\begin{picture}(0,0)
		\put(-249,128){\texttt{shift\_fw}}
		\put(-249,116){\texttt{shift\_bw}}
		\put(-49,60.5){\textcolor{white}{$3\times3$}}
		\put(-56.5,48.5){$sigmoid$}
		\put(-48,74){\texttt{pool}}
		\put(-55,85.2){\texttt{concat}}
		\put(-44,169){$tanh$}
		\put(-54,181){\textcolor{white}{$3\times3\times3$}}
		\put(-160,181.5){\textcolor{white}{C2D}}
		\put(-68,210){$X_1$}
		\put(-68,140){$R_1$}
		\put(-89,162){$Y_1$}
		\put(-53,107){$R_1$}
		\put(-53,5){$Z_1$}
		\put(-163,145){$X$}
		\put(-193,134){$Y_1$}
		\put(-193,107){$Y_2$}
		\put(-105,80){$Y_{1_{shift}}$}
		\put(-52,31){{\tiny$F_1$}}
		\put(-34,31){{\tiny$1$-$F_1$}}
		\put(-190,90){[$R_1$; $R_2$]}
		\put(-250,48.5){[$Y_{1_{shift}}$; $Y_{2_{shift}}$]}
		\put(-169,23){[$Z_1$; $Z_2$]}
		\vspace{-0.4cm}
	\end{picture}
	\caption{\ac{gsf} implementation with group gating, forward-backward temporal shift, and fusion. A gate is a single 3D convolution kernel with tanh calibration while fusion consists of a single 2D convolution kernel with sigmoid calibration. Thus very few parameters are added when \ac{gsf} is used to turn a C2D base model into a spatio-temporal feature extractor.
	}
	\label{fig:gsf}
\end{figure}

\newtext{Several 3D kernel factorization approaches have been developed for efficient video recognition. Fig.~\ref{fig:comp} illustrates the network schematics of some of these approaches.}
S3D, or R(2+1)D, P3D, decompose 3D convolutions into 2D spatial plus 1D temporal convolutions. \acs{tsm} replaces 1D temporal convolution with parameter-free \newtext{and zero FLOP} channel-wise temporal shift operations. GST uses group convolution where one group applies 2D spatial and the other 3D spatio-temporal convolution. GST furthermore applies point-wise convolution before and after the block to allow for interactions between spatial and spatio-temporal groups, and for channel reduction and up-sampling. In these modules, the feature flow is hard-wired by design, meaning that features are forwarded from one block to the next without data-dependent pooling, gating or routing decision. 

\ac{gsf} design, in Fig.~\ref{fig:comp}, is inspired by GST and TSM but replaces the hard-wired channel split \newtext{and concat aggregation} with a learnable spatial gating block \newtext{followed by a learnable fusion block to fuse the two splits of input features. The function of gate block is to selectively route the features through time-shifts to encode temporal information. The fuse block then merge the shifted (temporal) features with the residual (spatial) features. The gate and fuse blocks are learnable and perform gating and fusion operations in a data dependent manner.}

Based on the conceptual design in Fig.~\ref{fig:comp}, we instantiate \ac{gsf} as in Fig.~\ref{fig:gsf}. \newtext{We first describe the equations governing the implementation of \ac{gsf}}. Let $X$ be the $C\times T\times W\times H$ shaped input tensor to \ac{gsf}, where $C$ is the number of channels and $WH$, $T$ are the spatial and temporal dimensions, respectively. Let $X=[X_1,X_2]$ be the group=2 split of $X$ along the channel dimension, \newtext{$W_g=[W_{g_1}, W_{g_2}]$} be two \newtext{$1\times \nicefrac{C}{2}\times 3\times 3\times 3$} shaped {\em gating} kernels and \newtext{$W_f=[W_{f_1}$, $W_{f_2}]$ be two {\em fusion} kernels of shape $1\times2\times3\times3$.} Then, the \ac{gsf} output $Z = [Z_1, Z_2]$ is computed as

\begin{eqnarray}
Y_1 &=& tanh(W_{g_1} * X_1) \odot X_1 \label{eq:y1}\\
Y_2 &=& tanh(W_{g_2} * X_2) \odot X_2 \label{eq:y2}\\
R_1 &=& X_1-Y_1 \label{eq:r1}\\
R_2 &=& X_2-Y_2 \label{eq:r2}\\
Y_{1_{shift}} &=& \verb+shift_fw+(Y_1) \label{eq:y1_shift}\\
Y_{2_{shift}} &=& \verb+shift_bw+(Y_2) \label{eq:y2_shift}\\
F_1 &=& \sigma(W_{f_1} * [\phi(Y_{1_{shift}}; R_1)]) \label{eq:f1} \\
F_2 &=& \sigma(W_{f_2} * [\phi(Y_{2_{shift}}; R_2)]) \label{eq:f2} \\
Z_1 &=& F_1 \odot Y_{1_{shift }} + (1-F_1)\odot R_1 \label{eq:z1}\\
Z_2 &=& F_2 \odot Y_{2_{shift }} + (1-F_2)\odot R_2 \label{eq:z2}
\end{eqnarray}
where `$*$' represents convolution, `$\odot$' is Hadamard product, \texttt{shift\_fw}, \texttt{shift\_bw} is forward, backward temporal shift, \newtext{$\sigma$ is sigmoid function, [. ; .] is concatenation operation and $\phi$ is spatial average pooling.}

\newtext{The output features $X$ obtained from the spatial convolution, inherited from the 2D~\ac{cnn} base model, is applied to the gating module. The gating module splits the features into two groups across the channel dimension and generates two gating planes of shape $1\times T\times W\times H$, each for the two groups. The gating planes are then applied as spatial weight maps to each of the two feature groups to generate a pair of gated features and their residuals. The gated features are then group-shifted forward and backward in time to inject temporal information. The group-shifted features and their corresponding residuals are then applied to the fusion module. The fusion module generates a channel weight map of shape $C\times T\times1\times1$ for each of the two groups. The shifted features and their residuals are then fused via weighted averaging using the channel weight maps. This way, \ac{gsf} selectively mixes spatial and temporal information through learnable spatial gating and fusion mechanisms.} 

\noindent\textbf{Gating.}
Gating is implemented with a single spatio-temporal 3D kernel and tanh activation. With a 3D kernel we utilize short-range spatio-temporal information in the gating. $tanh$ provides spatial gating planes with values in the range $(-1,+1)$ and is motivated as follows. When the gating value at a feature location is 0 and that of the time-shifted feature was +1, then a \newtext{weighted} temporal feature averaging is performed at that location. If the gating value of the time-shifted feature was -1 instead, then a \newtext{weighted} temporal feature differencing is performed. Using $tanh$, the gating can thus learn to apply either of the two modes, location-wise.

\noindent\textbf{Fusion.} \newtext{The simplest choice for fusing the time-shifted temporal features with the residual is summation, as done in our preliminary work~\ac{gsm}~\cite{gsm}. However, since different channels of a feature tensor encode different level of information, fusion by summation may be sub-optimal. Instead ~\ac{gsf} performs weighted averaging on the two feature groups. For this, the group-shifted features and their corresponding residuals are concatenated followed by spatial average pooling operation, resulting in a tensor of shape $2\times C\times T$. Then a 2D convolution is applied on this tensor followed by $sigmoid$ non-linearity to obtain a weight map with shape $1\times C\times T$. This weight map is then reshaped to $C\times T \times 1\times1$ and is used to fuse the time-shifted features and their residuals via weighted averaging. When the fusion weight at a feature channel is 0, only the residual features with the spatial information is retained for further processing. On the other hand, if the channel weight is 1, the network will discard the residual spatial features and propagate the time-shifted temporal features. With other weight values, the model mixes the two feature groups selectively by mixing the channels.}

\noindent\textbf{Parameter count, complexity.} \newtext{The parameter count in the gating is $2\times (27\cdot \nicefrac{C}{2}) = 27\cdot C$ and that of the fusion module is $2\times(2\cdot3\cdot3) = 36$. The total parameter overhead introduced by \ac{gsf} is $27\cdot C + 36\approx27\cdot C$ since $C\gg36$ in practice. On the other hand, a typical C3D block such as the $3\times1\times1$ used in S3D~\cite{s3d} (Fig.~\ref{fig:comp}) has a parameter count of $C_{out}\cdot C\cdot3$ where $C_{out}\ge C$. Thus the parameter overhead introduced by GSF is far less than that of a C3D block since $27\ll 3\cdot C_{out}$.}

\newtext{The complexity of gating module is $H\cdot W\cdot T\cdot 27\cdot C$ while that of the fusion module is $C\cdot T\cdot 36$. The total complexity of \ac{gsf} is $H\cdot W\cdot T\cdot 27\cdot C + C\cdot T\cdot 18\approx H\cdot W\cdot T\cdot 27\cdot C$. For a C3D block performing 1D temporal convolution, the complexity is $H\cdot W\cdot T\cdot 3\cdot C\cdot C_{out}$. Since $27\ll 3\cdot C_{out}$, one can see that the complexity introduced by \ac{gsf} is far less than that of a 1D temporal convolution layer. Thus \ac{gsf} is far more efficient in terms of parameters and computations compared to the temporal convolution used in efficient implementations such as S3D or R(2+1)D.}

\subsection{Gate-Shift\newtext{-Fuse Networks}}
\label{sec:gsn}

\begin{figure}[t]
	\begin{subfigure}[b]{0.45\columnwidth}
		\centering
		\includegraphics[scale=0.1]{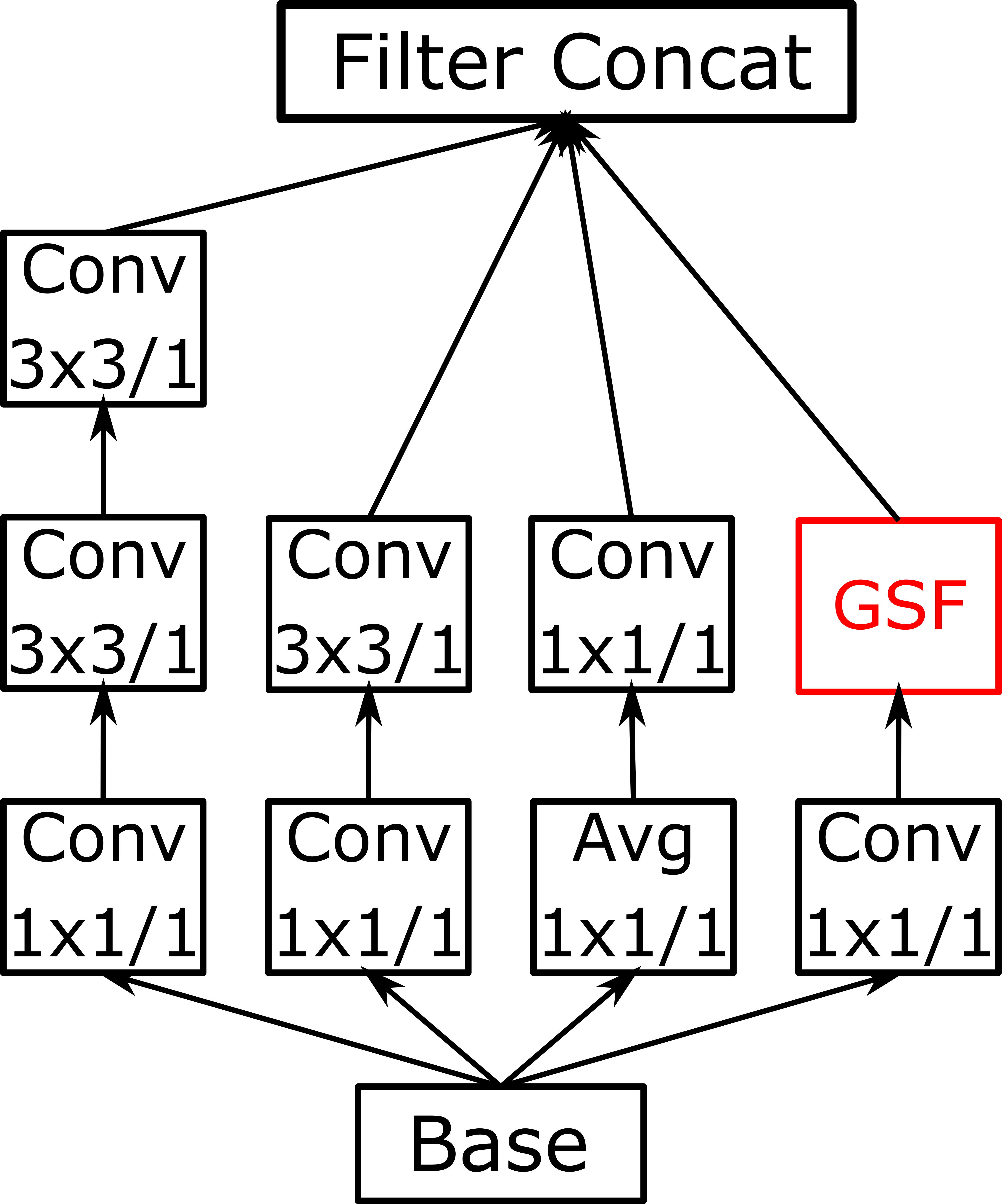}
	\end{subfigure} \hfill
	\begin{subfigure}[b]{0.45\columnwidth}
		\centering
		\includegraphics[scale=0.1]{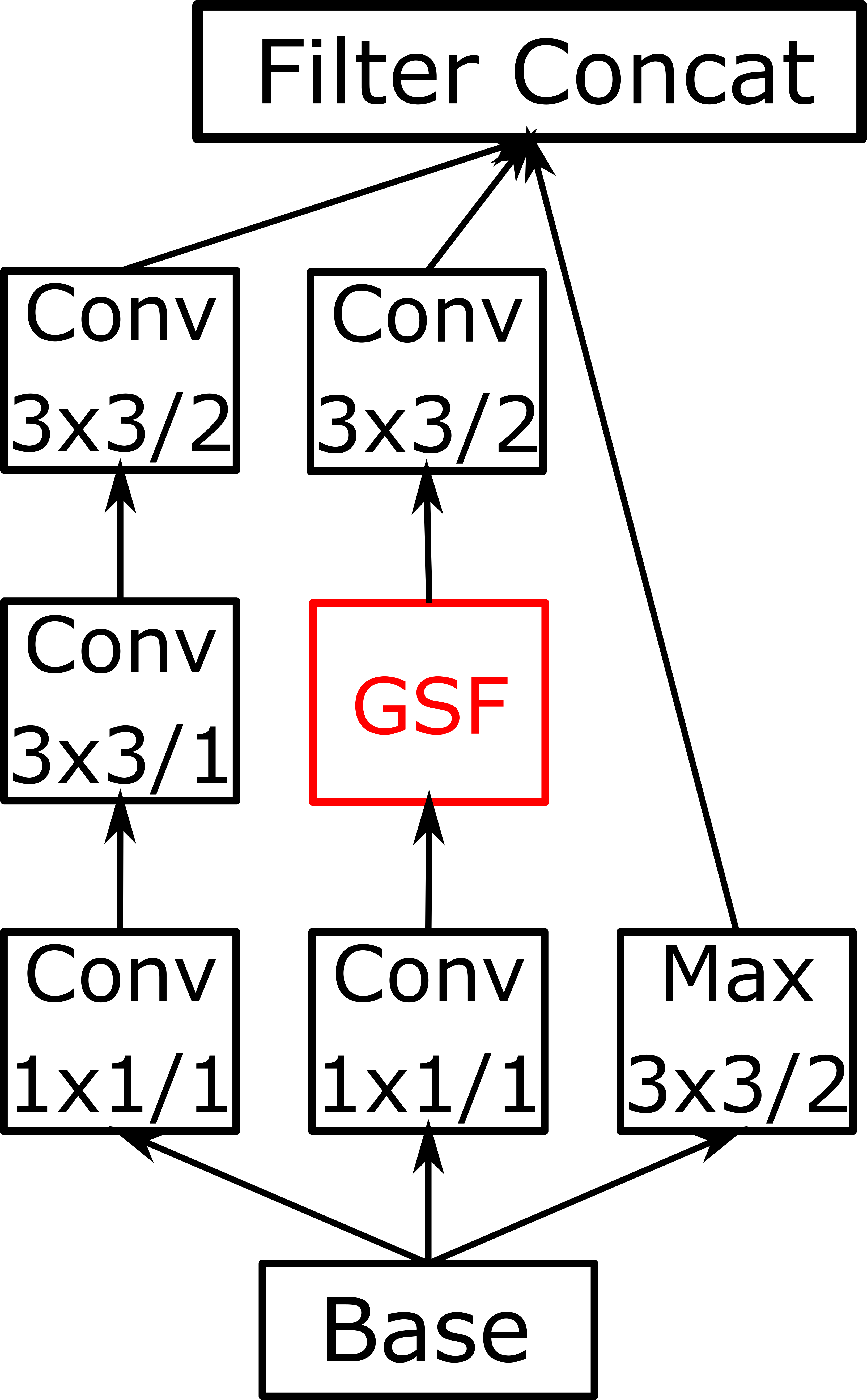}
	\end{subfigure}
	\caption{BN-Inception blocks with GSF. Kernel size and stride of conv and pool layers are annotated inside each block.
	}
	\label{fig:inc_blocks}
\end{figure}

\newtext{\ac{gsf} can be plugged into any 2D~\ac{cnn} architecture to inject temporal information into the features extracted by the backbone~\ac{cnn}. In this work, we choose two widely adopted 2D~\ac{cnn} families, Inception and ResNet, as the backbone~\acp{cnn}. As mentioned in Sec.~\ref{sec:gsf_module}, \ac{gsf} can be inserted into any layer in the backbone \ac{cnn} after a 2D convolution. For the Inception family, \ac{gsf} is applied to the least parametrized branch, \ie, the branch with the least number of convolutions~(cf.~Fig.\ref{fig:inc_blocks}). For ResNet based \ac{cnn}, \ac{gsf} is applied after the second convolution layer in the bottle neck layer of the backbone~(cf.~Fig.~\ref{fig:resnet_place}). Sec.~\ref{sec:model_dev} presents the experiments conducted to determine the optimum configuration for applying \ac{gsf} to the two \ac{cnn} families considered in this work.}

\newtext{We train the backbone \ac{cnn} with \ac{gsf} using the TSN framework. A fixed number of frames are uniformly sampled from the video and applied to the network. The network outputs a class prediction for each of the frames. The frame level predictions are then averaged to obtain the video level prediction. Training is done using cross-entropy loss computed from the video level prediction and the ground truth video label.}

		\section{Experiments and Results}
		\label{sec:experiments}
		This section presents an extensive set of experiments to evaluate \ac{gsf}.

\subsection{Datasets}
\label{sec:datasets}

We evaluate \acf{gsf} on five standard action recognition benchmarks, Something Something~\cite{goyal2017something} (SS-V1 and SS-V2), Kinetics 400~\cite{kin400}, Diving48~\cite{diving} and EPIC-Kitchens-100~\cite{epic100}.  Something-V1 and Something-V2 consists of fine-grained object manipulation actions comprising of 174 categories. Something-V1 consists of ~100K videos while Something-V2 is an updated version of Something-V1 with ~220K videos. Kinetics 400 is a large scale dataset consisting of 400 action categories with at least 400 clips per action class. Diving48 dataset contains around 18K videos with 48 fine-grained diving categories. EPIC-Kitchens-100 is the largest egocentric action recognition dataset with ~90K video clips. Each clip in the dataset is labelled with a verb and noun class from a possible set of 97 verb classes and 300 noun classes. The verb and noun category pairs are combined to form an action class.

All the datasets considered for evaluating \ac{gsf} differ considerably in the type of actions present in the videos. Something Something datasets consist of actions that are object agnostic. Thus understanding the actions carried out in the videos require strong temporal reasoning. On the other hand, the videos in EPIC-Kitchens-100 consists of actions that are object specific and requires encoding of the spatio-temporal patterns in the videos for recognizing an action. Both Kinetics 400 and Diving48 datasets contain human actions captured from a third person perspective and require understanding of the temporal dynamics of the human body in the video. While the background of the videos in Kinetics 400 can reveal useful information regarding the action category, the videos in Diving48 are captured with a uniform background thereby requiring encoding of the temporal progression of the actions.

\subsection{Implementation Details}
\label{sec:implementation_details}

\noindent \textbf{Training.} We train the models end-to-end using \ac{sgd} with momentum 0.9, weight decay $10^{-4}$ and an initial learning rate of 0.01, for 60 epochs. We use a cosine learning rate schedule with warmup (10 epochs) for adjusting the learning rate. The batch size used is 32 and a dropout of 0.5 is applied to avoid overfitting. 8 or 16 frames sampled from each video is used for predicting the action present in the video. Random scaling, cropping and flipping along with temporal jittering as presented in \cite{tsn} are used as data augmentations during training. The backbone \acp{cnn} are pretrained on ImageNet. Unless otherwise specified, we use the same setting for training \acp{gsf} on all the datasets used in this work.

\noindent \textbf{Testing.} Similar to contemporary works, we follow two approaches for evaluation: efficiency protocol and accuracy protocol. In the efficiency protocol, a single clip consisting of 8 or 16 frames are uniformly sampled from the video with each frame obtained by cropping the center portion of the video frame. In the accuracy protocol, we sample two clips for all the dataset except for Kinetics 400 from each video and extract three different crops from each frame. Thus a total of 6 clips are sampled from the input video. For Kinetics 400, we sample 10 clips and 3 crops from each video. The predictions of the individual clips are averaged to obtain the final prediction for the video.

\revnew{}{Source code of our implementation can be found online at \url{https://github.com/swathikirans/GSF}.}

\subsection{Model Development}
\label{sec:model_dev}

The model development is performed using SS-V1 dataset. We use 8 frames sampled from the video as input to the models and use the efficiency protocol for evaluating the testing performance.

\begin{table}[t]
	\centering
	\begin{tabular}{|c|c|c|c|}
		\hline
		\textbf{Branch} & \textbf{Accuracy (\%)} & \textbf{Params.} & \textbf{FLOPs} \\ \hline \hline
		Branch 1 & \revnew{45.49}{47.92} & 10.48M & 16.51G \\ \hline
		Branch 2 & \revnew{47.15}{47.97}  & 10.49M & 16.52G \\ \hline
		Branch 3 & \revnew{46.87}{48.05}  & 10.5M & 16.5G \\ \hline
		\textbf{Branch 4}  & \textbf{\revnew{48.09}{48.44}} & \textbf{10.5M} & \textbf{16.53G} \\ \hline
		All branches & \revnew{45.61}{46.55}  & 10.62M & 16.72G \\ \hline
	\end{tabular}
	\caption{Model design study done to determine the Inception branch that is most suitable for plugging in GSF.}
	\label{tab:bninception_place}
\end{table}

\begin{table}[h!]\small\centering\setlength\tabcolsep{4.5pt}
	\begin{tabular}{|c|c|c|c|}
		\hline
		\textbf{\% of channels shifted} & \textbf{Accuracy (\%)} & \textbf{Params.} & \textbf{FLOPs}\\ \hline \hline
		12.5 & \revnew{45.08}{45.62} & 10.46M & 16.45G\\ \hline
		25 & \revnew{45.44}{47.09} & 10.46M & 16.46G\\ \hline
		50 & \revnew{46.21}{47.65} & 10.47M & 16.48G\\ \hline
		75 & \revnew{47.02}{48.14} & 10.49M & 16.51G\\ \hline
		\textbf{100} & \textbf{\revnew{48.09}{48.44}} & \textbf{10.5M} & \textbf{16.53G}\\ \hline
	\end{tabular}
	\caption{Model design study on GSF-BNInception to analyze recognition accuracy vs number of channels shifted}
	\label{tab:bninception_ch}
\end{table}

\noindent \textbf{Inception backbones.} We use BNInception and InceptionV3 as the two \ac{cnn} backbones from the Inception family of models. Since the architectural designs are similar for the two models, we choose BNInception for identifying the optimum model configuration. 

Inception models have a multi-branch architecture as shown in Fig.~\ref{fig:inc_blocks}. We first perform analysis to identify which branch is more suitable for \ac{gsf} to be plugged into. The results of this analysis are provided in Tab.~\ref{tab:bninception_place}. From the table, one can see that \ac{gsf} performs the best when \ac{gsf} is plugged into the branch with the least number of convolution layers. 

\newtext{As mentioned in Sec.~\ref{sec:gsf_module}, temporal information is injected into the features by channel shifting. One can control the amount of temporal information by choosing the number of channels shifted. Shifting all the channels will allow the model to encode more temporal information. However, this is achieved at the expense of spatial information present in the features. There should be an optimum balance between the two.} In Tab.~\ref{tab:bninception_ch} we report the performance obtained by varying the number of channels shifted in each layers of the BNInception backbone. It can be seen from the table that shifting all the channels results in the best performance.

We use the configuration found from these two experiments for constructing \acp{gsf} with BNInception and InceptionV3 backbones.

\begin{figure}[t]
	\centering
	\includegraphics[scale=0.5]{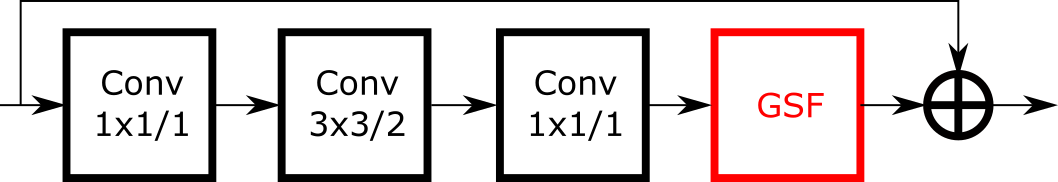}
	\caption{ResNet bottleneck layer with GSF. The kernel size and stride of conv layers are annotated inside each block.}
	\label{fig:resnet_place}
\end{figure}

\begin{table}[t]
	\centering
	\begin{tabular}{|c|c|c|c|}
		\hline
		\textbf{Branch} & \textbf{Accuracy (\%)} & \textbf{Params.} & \textbf{FLOPs}\\ \hline \hline
		After conv1 & \revnew{43.65}{46.95} & 23.97M & 33.53G \\ \hline
		After conv2 & \revnew{43.67}{47.15} & 23.97M & 33.4G \\ \hline
		\textbf{After conv3} & \textbf{\revnew{46.26}{47.63}} & \textbf{24.3M} & \textbf{34.43G} \\ \hline
		After all conv & \revnew{43.47}{46.26} & 24.52M & 35.25G \\ \hline
	\end{tabular}
	\caption{Model design analysis done to determine where to plug in GSF on ResNet bottleneck layers.}
	\label{tab:resnet_place}
\end{table}

\begin{table}[h!]\small\centering\setlength\tabcolsep{4.5pt}
	\begin{tabular}{|c|c|c|c|}
		\hline
		\textbf{\% of channels shifted} & \textbf{Accuracy (\%)} & \textbf{Params.} & \textbf{FLOPs}\\ \hline \hline
		12.5 & \revnew{47.5}{49.17} & 23.92M & 33.23G\\ \hline
		\textbf{25} & \textbf{\revnew{48.36}{50.17}} & \textbf{23.97M} & \textbf{33.4G}\\ \hline
		50 & \revnew{47.61}{50.08} & 24.08M & 33.75G\\ \hline
		75 & \revnew{47.27}{49.63} & 24.19M & 34.09G\\ \hline
		100 & \revnew{46.26}{47.63} & 24.3M & 34.43G\\ \hline
	\end{tabular}
	\caption{Model design study on GSF-ResNet-50 to analyze recognition accuracy vs number of channels shifted}
	\label{tab:resnet_ch}
\end{table}

\noindent \textbf{ResNet backbones.} The block diagram of the bottleneck layer of ResNet based \acp{cnn} are shown in Fig.~\ref{fig:resnet_place}. Tab.~\ref{tab:resnet_place} reports the performance of different configurations obtained by applying \ac{gsf} after different convolution layers present in the block of ResNet50. Following the results from Tab.~\ref{tab:resnet_place} we apply \ac{gsf} to after the second convolution of the ResNet block as shown in Fig.~\ref{fig:resnet_place}.

Similar to Inception architectures, we also conduct analysis on ResNet by shifting different number of channels. The results are reported in Tab.~\ref{tab:resnet_ch}. Unlike with the case of Inception models, the best performance is obtained when only 25\% of the channels are shifted. This can be attributed to the fact that shifting all the channels will affect the spatial modelling capability of the network. In the case of Inception based models, the branches without \ac{gsf} will encode the spatial information. Hence, \ac{gsf} can be applied to all the channels in the corresponding branch to inject temporal information to the features.

\subsection{Ablation Analysis}
\label{sec:ablation}
In this section, we report the ablation analysis performed on the validation set of Something-V1 dataset. In all the experiments, we apply 8 frames as input to the network and report accuracy using the efficiency protocol.

\begin{table}[t]\small\centering\setlength\tabcolsep{4.5pt}
	\begin{tabular}{|c|c|c|c|}
		\hline
		\textbf{Gating} & \textbf{Fusion} & Model & \textbf{Accuracy (\%)} \\ \hline \hline
		Learnable & Learnable & GSF & \revnew{48.09}{48.44} \\ \hline
		Learnbale & Sum & GSM & 47.24 \\ \hline
		1 & Sum & TSM & 45.57\\ \hline
		-1 & Sum &  TDN & 45.89 \\ \hline
		0 & Sum & TSN & 17.25 \\ \hline		
	\end{tabular}
	\caption{Ablation study on GSF}
	\label{tab:gsf_ablation}
\end{table}

We first ablate the various components of \ac{gsf} and validate the design choices. We choose BNInception as the backbone model due to its reduced complexity. The results are reported in Tab.~\ref{tab:gsf_ablation}. By removing the channel fusion, \ac{gsf} converges to \acs{gsm}~\cite{gsm}, our prior work on top of which \ac{gsf} is built upon, and underperforms by \revnew{$0.85\%$}{$1.2\%$}. This validates the importance of the learned channel fusion. We then fix the spatial gating to a fixed value of 1 which results in an accuracy of $45.57\%$. This is similar to \ac{tsm}~\cite{tsm} where temporal shifting is applied to all the channels. With a fixed gating of -1, the networks performs differencing operation between the adjacent frames and results in $45.89\%$. This is a special case of \ac{tdn}~\cite{tdn} without the image subnetwork. Finally, we fix the gating to 0 which converges the model to the baseline, ~\acs{tsn}, without any temporal interaction between the frame level features, and drops the accuracy by \revnew{$30\%$}{$31\%$}.

\begin{table}[t]\small\centering\setlength\tabcolsep{4.5pt}
	\begin{tabular}{|c|c|c|c|}
		\hline
		\textbf{Model} & \textbf{Accuracy (\%)} & \textbf{Params.} & \textbf{FLOPs}\\ \hline \hline
		BNInception (baseline) & 17.25 & 10.45M & 16.43G\\ \hline
		BNInception + 1 GSF & \revnew{30.39}{31.9} & 10.46M & 16.44G\\ \hline
		BNInception + 5 GSF & \revnew{42.43}{45.24} & 10.48M & 16.46G\\ \hline
		BNInception + 10 GSF & \revnew{48.09}{48.44} & 10.5M & 16.53G\\ \hline
	\end{tabular}
	\caption{Recognition Accuracy by varying the number of GSF added to BNInception backbone.}
	\label{tab:ablation_gsfcount}
\end{table}

\begin{table}[h!]\small\centering\setlength\tabcolsep{4.5pt}
	\begin{tabular}{|c|c|c|c|}
		\hline
		\textbf{Model} & \textbf{Accuracy (\%)} & \textbf{Params.} & \textbf{FLOPs}\\ \hline \hline
		ResNet-50 (baseline) & 17.22 & 23.86M & 33.06G\\ \hline
		ResNet-50 + 4 GSF & \revnew{44.71}{45.97} & 23.92 & 33.09G\\ \hline
		ResNet-50 + 8 GSF & \revnew{47.28}{48.54} & 23.95M & 33.14G\\ \hline
		ResNet-50 + 16 GSF & \revnew{48.36}{50.17} & 23.97M & 33.4G\\ \hline
	\end{tabular}
	\caption{Recognition Accuracy by varying the number of GSF added to ResNet50 backbone.}
	\label{tab:ablation_gsfcount_res50}
\end{table}

Next we compare the performance improvement obtained by adding different number of \ac{gsf} to the backbone \ac{cnn}. Tab.~\ref{tab:ablation_gsfcount} shows the results of this experiment conducted on BNInception backbone. Baseline is the standard \ac{tsn} architecture, with an accuracy of $17.25\%$. We then applied \ac{gsf} at the last Inception block of the \ac{cnn}. This improved the recognition performance by \revnew{$13\%$}{$14.65\%$}. By adding \ac{gsf} to the last 5 layers of the backbone, the performance improves by \revnew{$25\%$}{$27.99\%$}. Thus, a consistent improvement in performance can be observed by plugging in \ac{gsf} to the backbone layers of the \ac{cnn}. With \ac{gsf} added to all the Inception blocks in the \ac{cnn}, an absolute improvement of \revnew{$30\%$}{$31\%$} is obtained, over the baseline, to reach an accuracy of \revnew{$48.09\%$}{$48.44\%$}. This huge improvement is obtained only with an overhead of $0.48\%$ and $0.55\%$ in terms of parameters and complexity, respectively.

We also conducted the above experiment on ResNet-50 backbone. The GSF follows the setting mentioned in Sec.~\ref{sec:model_dev} and the results are reported in Tab.~\ref{tab:ablation_gsfcount_res50}. By plugging in \ac{gsf} to the last four bottleneck layers of the backbone, a significant improvement of \revnew{$+27\%$}{$28\%$} in recognition accuracy is obtained. This is achieved only with a very small overhead in number of parameters and computations. As observed in the case of BNInception backbone, the recognition performance increased consistently by plugging in \ac{gsf} to all the bottleneck layers of the backbone \ac{cnn}.

\subsection{State-of-the-Art Comparison}
\label{sec:sota_comp}

\begin{table*}[t]
	\centering
	\begin{tabular}{|c|c|c|c|c|c|c|}
		\hline
		\multirow{2}{*}{\textbf{Method}} & \multirow{2}{*}{\textbf{Backbone}} &  \multirow{2}{*}{\textbf{\#Frames}} & \multirow{2}{*}{\textbf{Params (M)}} & \multirow{2}{*}{\textbf{GFLOPs}} & \multicolumn{2}{c|}{\textbf{Accuracy (\%)}}   \\ \cline{6-7}
		& & & & & SS-V1 & SS-V2\\ \hline \hline 
		\acs{tsn}~\cite{tsn}  & BN-Inception & $16\times1\times1$ & 10.45 & $32.87\times1\times1$ &  17.44 & 32.71 \\ \hline
		
		MultiScale \acs{trn}~\cite{trn}  & BN-Inception & $8\times$NA & NA & NA & 34.44 & 48.8 \\ \hline
		MFNet~\cite{mfnet}  & ResNet-101 & $10\times1\times1$ & NA &  NA & 43.92 & - \\ \hline
		
		S3D-G~\cite{s3d} & InceptionV1 & $64\times1\times1$ & 11.6 & $71.38\times1\times1$ & 48.2 & - \\ \hline
		
		TrajectoryNet~\cite{trajectoryNet}  & ResNet-18 &  16$\times$NA & NA & NA & 47.8 & - \\ \hline
		
		ABM~\cite{abm}  & ResNet-50 & $16\times3\times1$ & NA & NA & 46.81 & 61.25 \\ \hline
		\acs{tsm}~\cite{tsm}  & ResNet-50 & $16\times1\times1$ &  24.3 & $65\times1\times1$ & 47.3 & 61.2 \\ \hline
		
		\revnew{}{RubiksNet}~\cite{rubiksnet} & - & \revnew{}{$8\times1\times 1$} & \revnew{}{8.5} & \revnew{}{$15.8\times1\times 1$} & 46.4 & \revnew{}{59.0} \\ \hline
		
		STM~\cite{stm}  & ResNet-50 & $16\times3\times10$ & 24 & $66.5\times3\times10$ & 50.7 & 64.2 \\ \hline
		
		bLVNet-TAM~\cite{blvnet-tam} & bLResNet-50 & $16\times1\times2$ & 25 & $47\times1\times2$ & 48.4 & 61.7\\ \hline
		bLVNet-TAM~\cite{blvnet-tam} & bLResNet-101 & $32\times1\times2$ & 40.2 & $128.6\times1\times2$ & 53.1 & 65.2\\ \hline
		CorrNet~\cite{corrnet} & ResNet-101 & 32$\times1\times10$ & NA & $224\times1\times10$ & 51.1 & - \\ \hline 		
	
		TPN~\cite{tpn} & ResNet-50 & $8\times10\times1$ & NA & NA & 49.0 & 62 \\ \hline 
		
		TEA~\cite{tea} & ResNet-50 & $8\times3\times10$ & NA & $35\times3\times10$ & 51.7 & NA  \\ \hline
		TEA~\cite{tea} & ResNet-50 & 16$\times$3$\times$10 & NA & 70$\times$3$\times$10 & 52.3 & - \\ \hline
		
		MSNet~\cite{motionsqueeze} & ResNet-50 & $8\times1\times1$ & 24.6 & $34\times1\times1$ & 50.9 & 63 \\ \hline
		MSNet~\cite{motionsqueeze} & ResNet-50 & $16\times1\times1$ & 24.6 & $67\times1\times1$ & 52.1 & 64.7 \\ \hline 
		MVFNet~\cite{mvfnet} & ResNet-50 & $16\times3\times2$ & - & $66\times3\times2$ & 52.6 & 65.2 \\ \hline
		CT-Net~\cite{ctnet} & ResNet-50 & $16\times3\times2$ & - & $75\times3\times2$ & 53.4 & 65.9 \\ \hline \hline
		
		I3D~\cite{carreira2017quo}~from~\cite{wang2018videos}  & ResNet-50 & $32\times1\times2$ & 28 & $153\times1\times2$ & 41.6 & - \\ \hline
		Non-local~from~\cite{wang2018videos}  & ResNet-50 & $32\times1\times2$ & 35.3 & $168\times1\times2$ & 44.4 & - \\ \hline
		GCN+Non-local~\cite{wang2018videos}  & ResNet-50 & $32\times1\times2$ & 62.2 & $303\times1\times2$ & 46.1 & - \\ \hline
		\acs{gst}~\cite{gst}  & ResNet-50 & $16\times1\times1$ & 21 & $59\times1\times1$ & 48.6 & 62.6 \\ \hline
		Martinez \etal~\cite{martinez2019action}  & ResNet-50 & NA & NA & NA & 50.1 & - \\ \hline
		Martinez \etal~\cite{martinez2019action}  & ResNet-152 & NA & NA & NA & 53.4 & - \\ \hline  
		\revnew{}{MoViNet-A3}~\cite{movinet} & - & \revnew{}{$50\times1\times 1$} & \revnew{}{5.3} & \revnew{}{$23.7\times1\times 1$} & - & \revnew{}{64.1} \\ \hline\hline
	
		\multirow{14}{*}{\ac{gsf}} 
		& BNInception & \multirow{3}{*}{$8\times1\times1$} & 10.5 & $16.53\times1\times1$ & \revnew{48.09}{48.44}  & \revnew{61.02}{61.29} \\ \cline{2-2} \cline{4-7}
		& ResNet-50  &  & 23.97 & $33.4\times1\times1$ & \revnew{48.36}{50.17}  & \revnew{61.46}{62.19} \\ \cline{2-2} \cline{4-7}
		& InceptionV3  & & 22.22 & $26.96\times1\times1$ & \revnew{48.68}{51.52} & \revnew{62.64}{62.83} \\ \cline{2-7} 
		& BNInception & \multirow{3}{*}{$8\times3\times2$} & 10.5 & $16.53\times3\times2$ & \revnew{49.05}{50.17} & \revnew{62.51}{62.76} \\ \cline{2-2} \cline{4-7}
		& ResNet-50  &  & 23.97 & $33.4\times3\times2$ & \revnew{48.36}{51.53} & \revnew{61.46}{64.01} \\ \cline{2-2} \cline{4-7}
		& InceptionV3  &  & 22.22 & $26.85\times3\times2$ & \revnew{49.84}{52.53} & \revnew{64.7}{64.59} \\ \cline{2-7} 
		& BNInception & \multirow{4}{*}{$16\times1\times1$} & 10.5 & $33.06\times1\times1$ & \revnew{49.31}{50.63}  & \revnew{62.91}{63.4}\\ \cline{2-2} \cline{4-7}
		& ResNet-50  & & 23.97 & $66.8\times1\times1$ & \revnew{50.37}{51.54}  & \revnew{63.41}{63.97} \\ \cline{2-2} \cline{4-7}
		& InceptionV3  &  & 22.22 & $53.93\times1\times1$ & \revnew{51.13}{53.13}  & \revnew{63.88}{64.48} \\ \cline{2-2} \cline{4-7}
		& ResNet-50+MS & & 24.05 & $68.05\times1\times1$ & \revnew{52.59}{52.7} & \revnew{63.96}{64.54} \\ \cline{2-7} 
		& BNInception &  \multirow{4}{*}{$16\times3\times2$} & 10.5 & $33.06\times3\times2$ & \revnew{50.71}{51.51} & \revnew{64.35}{64.57} \\ \cline{2-2} \cline{4-7}
		& ResNet50  &  & 23.97 & $66.8\times3\times2$ & \revnew{51.42}{52.33} & \revnew{65.07}{65.23} \\  \cline{2-2} \cline{4-7}
		& InceptionV3  &  & 22.22 & $53.93\times3\times2$ & \revnew{52.1}{54.09} & \revnew{65.18}{65.60} \\ 
		\cline{2-2} \cline{4-7}
		& ResNet-50+MS & & 24.05 & $68.05\times3\times2$ & \revnew{53.78}{54.04} & \revnew{65.61}{65.73} \\ \hline
	\end{tabular}
	\caption{Comparison to \sota on Something-V1 and Something-V2 datasets.}
	\label{tab:ssv1_sota}
\end{table*}

\textbf{Something-Something.} Tab.~\ref{tab:ssv1_sota} compares the recognition performance of \ac{gsf} with \sota approaches on SS-V1 and SS-V2 datasets. For fair comparison, we compare only with those methods that use RGB frames for feature extraction. We also report the number of frames used by each approach along with the number of parameters and the computational complexity in terms of \acsp{flop} in the table. Since different approaches use different backbone \acp{cnn}, we also report the backbones used for fair comparison. The table is split into two, the first part lists approaches that use 2D~\ac{cnn} or efficient implementations of 3D~\ac{cnn} as backbone. The second part shows the methods that rely on heavier 3D~\acp{cnn} for spatio-temporal feature extraction. Our baseline TSN~\cite{tsn} obtains 17.52\% and 32.71\% recognition accuracy on SS-V1 and SS-V2 datasets, respectively. Applying \ac{gsf} to the backbone \ac{cnn}, an absolute gain of \revnew{$+31.79\%$}{$+33.19\%$} and \revnew{$+30.2\%$}{$+30.69\%$} is obtained on the two datasets. This huge improvement is obtained with an increase of only +0.48\% and +0.55\% on the number of parameters and \acsp{flop}. Compared to the top performing approaches that use 2D~\ac{cnn}, such as TEA~\cite{tea} and MSNet~\cite{motionsqueeze}, \ac{gsf} performs competitively. On the other hand, approaches that uses 3D~\ac{cnn} such as I3D~\cite{carreira2017quo} and Non-local networks~\cite{nonlocal}, considerably falls behind \ac{gsf}. The approach from~\cite{martinez2019action} uses a deeper 3D~\ac{cnn} (I3D ResNet-152). When compared to the performance obtained with a backbone ~\ac{cnn} of equivalent depth (ResNet-50), \ac{gsf} outperforms the approach from~\cite{martinez2019action} by \revnew{$+1.3\%$}{$+2.2\%$}. By plugging in the MotionSqueeze module~\cite{motionsqueeze} on the GSF version of ResNet-50, our model surpasses the recently developed \sota approaches MVF-Net~\cite{mvfnet} and CT-Net~\cite{ctnet}, with same backbone, achieving an accuracy of \revnew{$53.78\%$}{$54.04\%$} on SS-V1.

\noindent \textbf{Kinetics400.} We compare the performance of \ac{gsf} with \sota methods on Tab.\ref{tab:kin400_sota}. All two \ac{gsf} models are trained with 16 frames and the accuracy protocol with 10 clip sampling during inference. As done previously, we split the table into two sections: top section comparing methods that use 2D~\ac{cnn} backbones and bottom section lists the approaches that use 3D~\ac{cnn} backbone. As mentioned in Sec.~\ref{sec:datasets}, Kinetics400 consists of videos of short duration (~10s) that do not require strong temporal reasoning. This is reflected in Tab.~\ref{tab:kin400_sota}, where the approaches that use 3D~\ac{cnn} for feature extraction performs significantly better than those that use 2D~\acp{cnn}. Compared to approaches that use 2D~\ac{cnn}, the two \ac{gsf} models perform competitively with the InceptionV3 based model outperforming its ResNet-50 counterpart by \revnew{$0.5\%$}{$0.7\%$}.

\begin{table}[t]\small
	\centering
	\begin{tabular}{|c|c|c|}
		\hline
		\textbf{Method} & \textbf{Backbone} & \textbf{Accuracy (\%)} \\ \hline \hline
		bLVNet-TAM~\cite{blvnet-tam} & bLResNet-50 & 73.5\\ \hline
		STM~\cite{stm} & ResNet-50 & 73.7\\ \hline
		TSM~\cite{tsm} & ResNet-50 & 74.1 \\ \hline
		Martnez \etal~\cite{martinez2019action} & NA & 74.3 \\ \hline
		TEA~\cite{tea} & ResNet-50 & 76.1 \\ \hline
		MFNet~\cite{mfnet} & ResNet-50 & 76.4\\ \hline MVFNet~\cite{mvfnet} & ResNet-50 & 77.0\\ \hline
		CT-Net~\cite{ctnet} & ResNet-50 & 77.3\\ \hline \hline
		I3D ~\cite{mfnet} & ResNet-101 + NL & 77.7 \\ \hline
		Martinez \etal~\cite{martinez2019action} & NA & 78.8 \\ \hline
		ip-CSN~\cite{mfnet} & ResNet-152 & 79.2\\ \hline
		TPN~\cite{mfnet} & ResNet-101 & 78.9\\ \hline 
		SlowFast~\cite{mfnet} & ResNet-101 + NL & 79.8\\ \hline 
		
		X3D-XXL~\cite{mfnet} & - & 80.4\\ \hline 
		
		\revnew{}{MoViNet-A6}~\cite{movinet} & - & \revnew{}{81.5} \\ \hline \hline
		\multirow{2}{*}{\ac{gsf}} & ResNet-50 & \revnew{74.74}{74.81} \\ \cline{2-3}
		& InceptionV3 & \revnew{75.31}{75.57} \\ \hline
	\end{tabular}
	\caption{Comparison to \sota efficient architectures on Kinetics400.}
	\label{tab:kin400_sota}
\end{table}

\noindent \textbf{EPIC-Kitchens-100.} Comparison of \ac{gsf} with \sota approaches on the validation set of EPIC-Kitchens-100 is performed on Tab.~\ref{tab:epic100_sota}. For this dataset, we apply autoaugment~\cite{autoaugment} in addition to random cropping and random flipping as augmentations. All models are trained with 16 frames as a multitask problem for predicting verb, noun and action classes. Accuracy protocol is used during inference. InceptionV3 and ResNet-50 models are initialized with Kinetics400 pretrained weights while ResNet-101 backbone is initialized with ImageNet pretrained weights. From the table one can see that \ac{gsf} models outperform the other approaches significantly\revnew{}{, except for MoViNet~\cite{movinet} which is a NAS based approach using 3D convolutions}. One interesting finding is that in contrast to what was observed for Something and Kinetics datasets, the ResNet based backbones result in better performance than InceptionV3 based model.

\begin{table}[t]\small
	\centering
	\begin{tabular}{|c|c|c|c|c|}
		\hline
		\multirow{2}{*}{\textbf{Method}} & \multirow{2}{*}{\textbf{Backbone}}& \multicolumn{3}{c|}{\textbf{Accuracy (\%)}} \\ \cline{3-5}
		& &Verb & Noun & Action \\ \hline \hline
		TSN~\cite{tsn}$^*$ & ResNet-50 & 60.18 & 46.03 & 33.19 \\ \hline
		TRN~\cite{trn}$^*$ & ResNet-50 & 65.88 & 45.43 & 35.34 \\ \hline
		TSM~\cite{tsm}$^*$ & ResNet-50 & 67.86 & 49.01 & 38.27 \\ \hline
		SlowFast~\cite{slowfast}$^*$ & ResNet-50 & 65.56   & 50.02 & 38.54\\ \hline 
		\revnew{}{MoViNet-A6}~\cite{movinet} & - & \revnew{}{72.2} & \revnew{}{57.3} & \revnew{}{47.7}\\ \hline\hline
		\multirow{3}{*}{\acs{gsf}} & InceptionV3 & \revnew{68.89}{68.35} & \revnew{51.42}{52.71} & \revnew{43.11}{43.42} \\ \cline{2-5}
		& ResNet-50 &  \revnew{68.88}{68.76} & \revnew{52.73}{52.74} & \revnew{43.84}{44.04} \\ \cline{2-5}
		& ResNet-101 & \revnew{69.06}{69.97} & \revnew{53.18}{54.01} & \revnew{44.48}{44.78} \\ \hline
	\end{tabular}
	\caption{Comparison to \sota on the validation set of EPIC-Kitchens-100. $^*$: Results reported from~\cite{epic100}}
	\label{tab:epic100_sota}
\end{table}

\begin{table}[t]\small
	\centering
	\begin{tabular}{|c|c|}
		\hline
		\textbf{Method} & \textbf{Accuracy (\%)} \\ \hline \hline
		I3D~\cite{carreira2017quo}~from~\cite{tqn} & 48.3 \\ \hline
		S3D~\cite{s3d}~from~\cite{tqn} & 50.6 \\ \hline
		SlowFast~\cite{slowfast}~from~\cite{timesformer} & 77.6\\ \hline
		GST~\cite{gst}~from~\cite{tqn} & 78.9 \\ \hline
		TimeSformer-L~\cite{timesformer} & 81\\ \hline
		TQN~\cite{tqn} & 81.8 \\ \hline \hline
		\acs{gsf} & \revnew{89.75}{90.51} \\ \hline
	\end{tabular}
	\caption{Comparison to \sota on Diving48.}
	\label{tab:diving_sota}
\end{table}

\noindent \textbf{Diving48.} We compare the performance of Diving48 dataset on Tab.~\ref{tab:diving_sota}. We use 16 frames to train the \ac{gsf} model with InceptionV3 backbone and inference is performed following the accuracy protocol. All training settings are the same as mentioned in Sec.~\ref{sec:implementation_details} except for the batch size. For Diving48, we used a smaller batch size of 8 during training. From the table, one can see that \ac{gsf} model outperforms approaches that uses 3D~\ac{cnn} backbones such as I3D~\cite{carreira2017quo}, SlowFast~\cite{slowfast} and GST~\cite{gst} by a significant margin.

\begin{figure*}[h]
	\begin{subfigure}[b]{0.4\linewidth}
		\centering
		\includegraphics[height=0.7\columnwidth]{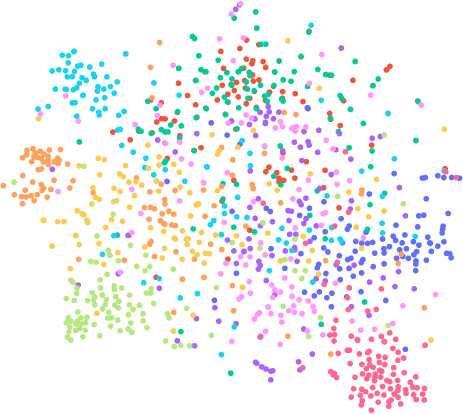}
		\caption{}
		\label{fig:tsn_group10_tsne}
	\end{subfigure}  \hskip -2mm
	\begin{subfigure}[b]{0.55\linewidth}
		\centering
		\includegraphics[height=0.59\columnwidth]{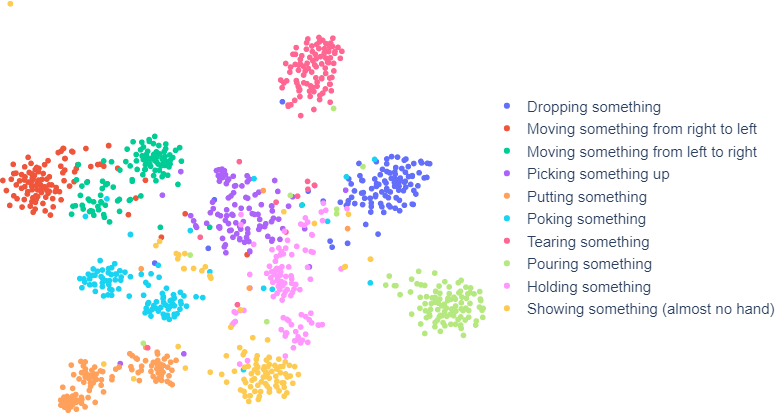}
		\caption{}
		\label{fig:gsf_group10_tsne}
	\end{subfigure}  \hfill
	\caption{t-SNE projection of output features obtained from the layer preceeding the final linear layer for (a) TSN baseline and (b) TSN baseline with GSF. Samples from the 10 action groups defined in~\cite{goyal2017something} are visualized.}
	\label{fig:tsne_group10}
\end{figure*}

\begin{figure*}[h]
	\centering
	\includegraphics[width=\linewidth]{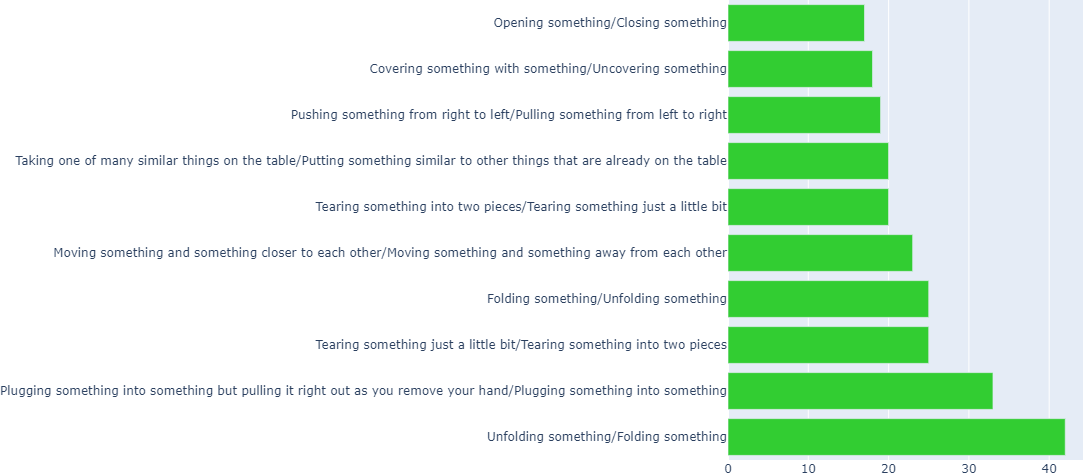}
	\caption{Action classes with the highest improvement over TSN baseline. X-axis shows the the number of corrected samples for each class. Y-axis labels are in the format true label (GSF)/predicted label (TSN).}
	\label{fig:improved_classes}
\end{figure*}

\subsection{Discussion}
We perform further analysis of \ac{gsf} in this section. The t-SNE plot of features obtained from the output of the layer preceding the final \ac{fc} layer of BNInception backbone is plotted in Fig.~\ref{fig:tsn_group10_tsne}. We visualize the 10 action groups defined in~\cite{goyal2017something} in the plot. From the figure, one can see better separation of the features when \ac{gsf} is plugged into the backbone \ac{cnn}. This allows the model to correctly recognize the action present in the video and results in improved recognition performance compared to the baseline \ac{tsn}.

\begin{figure*}[h!]
	\centering
	\begin{subfigure}[b]{0.45\textwidth}
		\centering
		\includegraphics[scale=0.22]{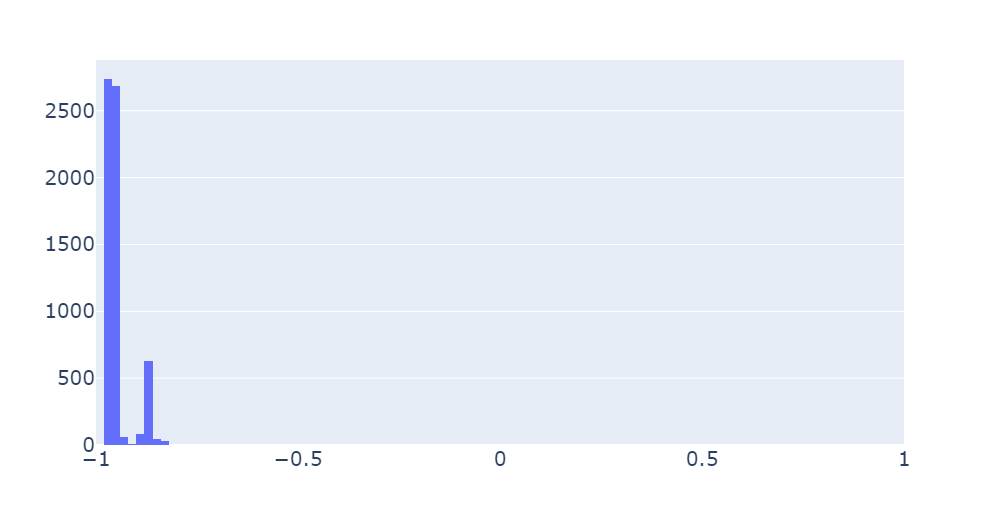}
		\caption{Layer 1}
	\end{subfigure}
	\begin{subfigure}[b]{0.45\textwidth}
		\centering
		\includegraphics[scale=0.22]{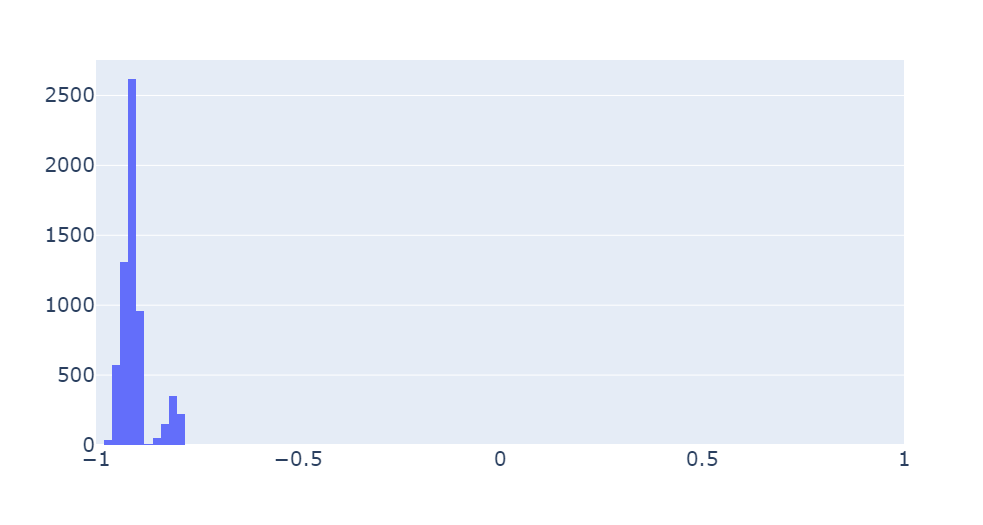}
		\caption{Layer 2}
	\end{subfigure}
	\begin{subfigure}[b]{0.45\textwidth}
		\centering
		\includegraphics[scale=0.22]{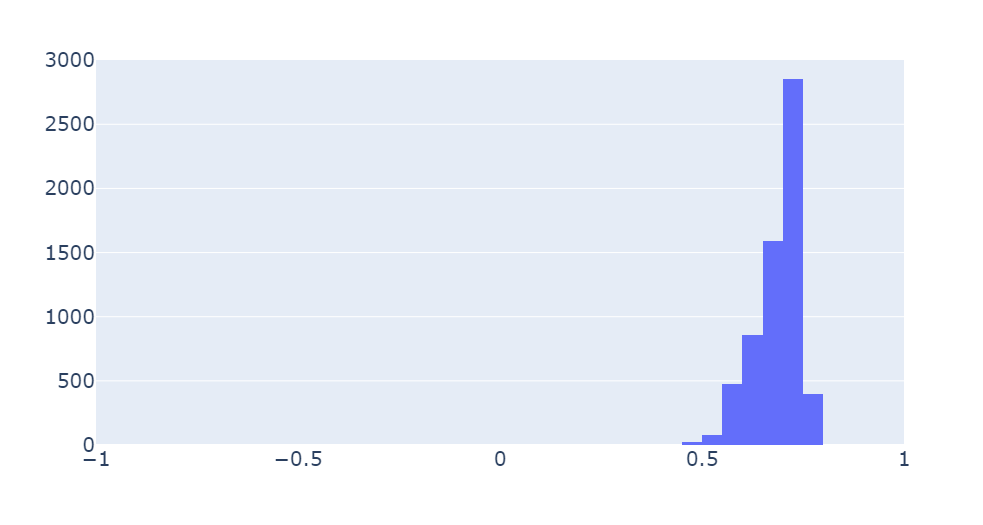}
		\caption{Layer 3}
	\end{subfigure}
	\begin{subfigure}[b]{0.45\textwidth}
		\centering
		\includegraphics[scale=0.22]{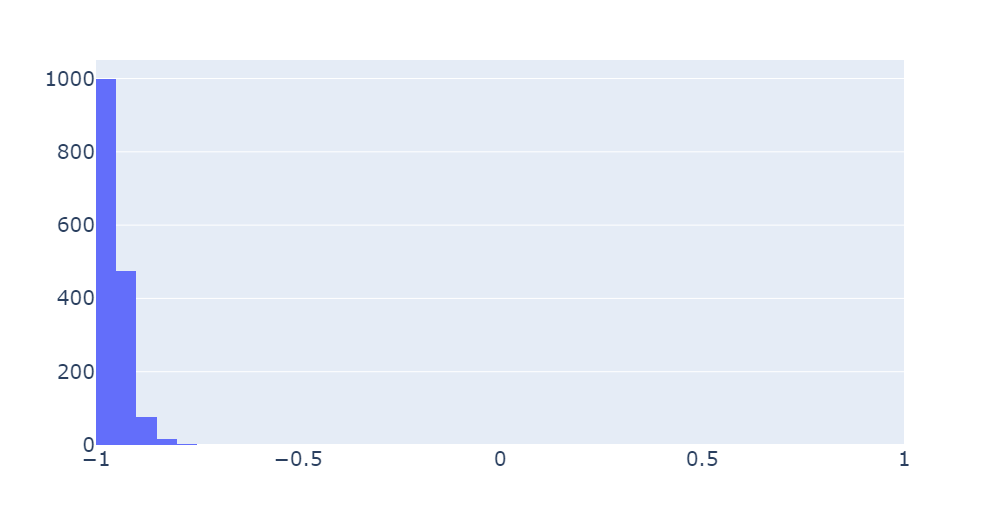}
		\caption{Layer 4}
	\end{subfigure}
	\begin{subfigure}[b]{0.45\textwidth}
		\centering
		\includegraphics[scale=0.22]{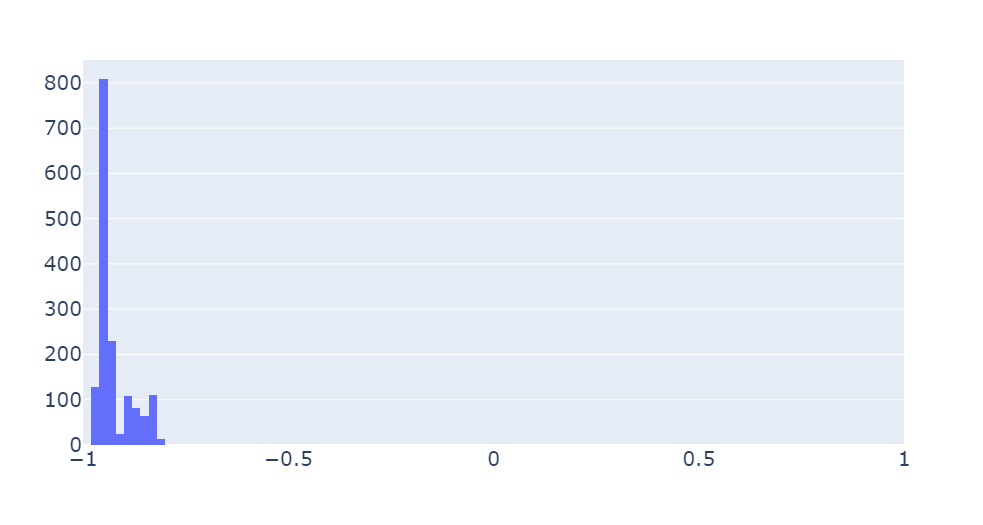}
		\caption{Layer 5}
	\end{subfigure}
	\begin{subfigure}[b]{0.45\textwidth}
	\centering
	\includegraphics[scale=0.22]{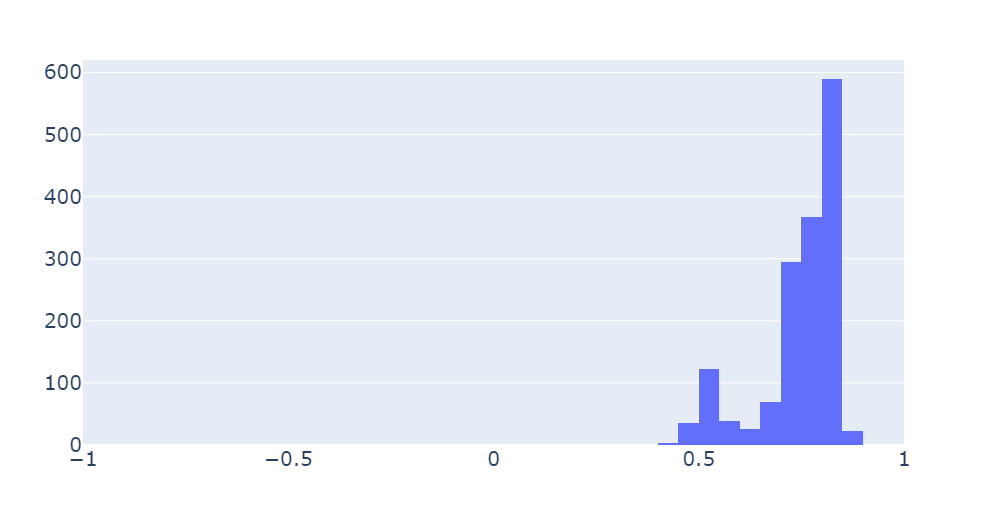}
	\caption{Layer 6}
	\end{subfigure}
		\begin{subfigure}[b]{0.45\textwidth}
		\centering
		\includegraphics[scale=0.22]{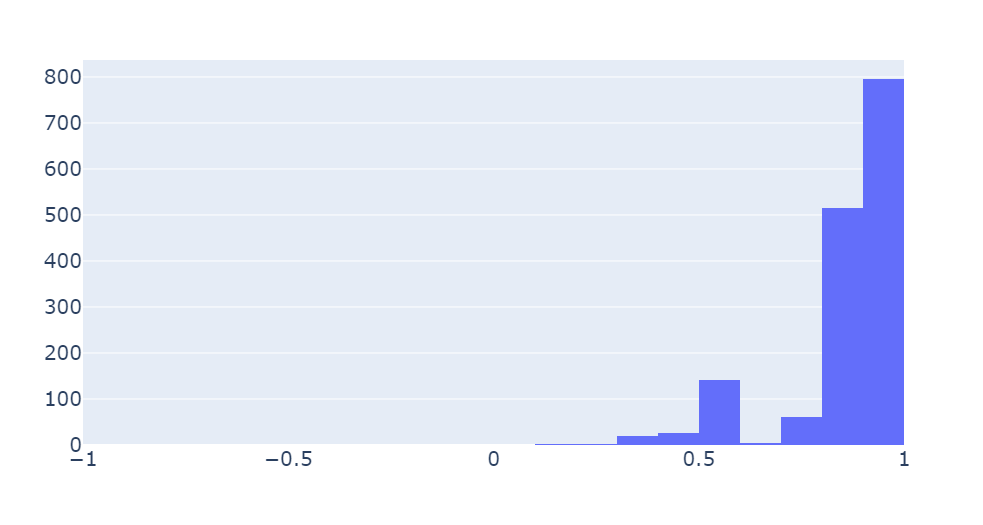}
		\caption{Layer 7}
	\end{subfigure}
		\begin{subfigure}[b]{0.45\textwidth}
		\centering
		\includegraphics[scale=0.22]{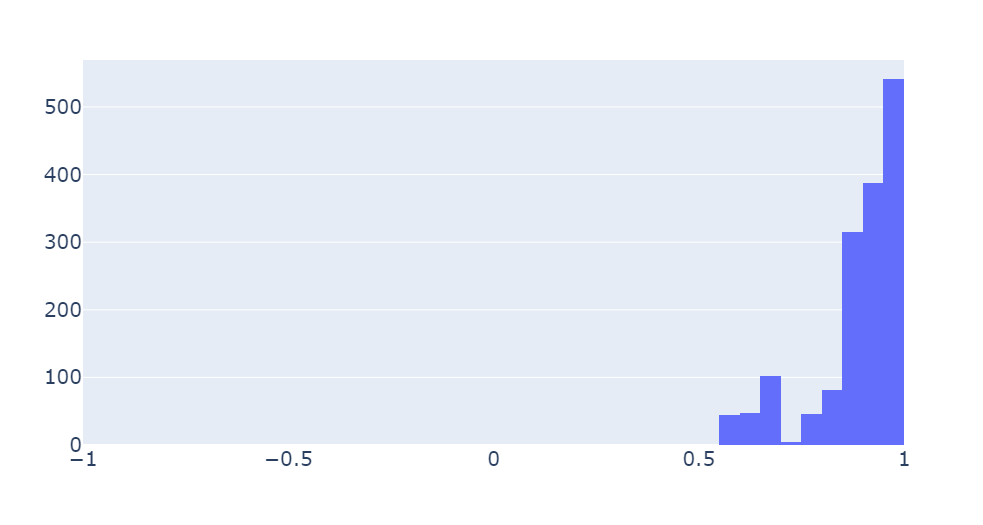}
		\caption{Layer 8}
	\end{subfigure}
		\begin{subfigure}[b]{0.45\textwidth}
		\centering
		\includegraphics[scale=0.22]{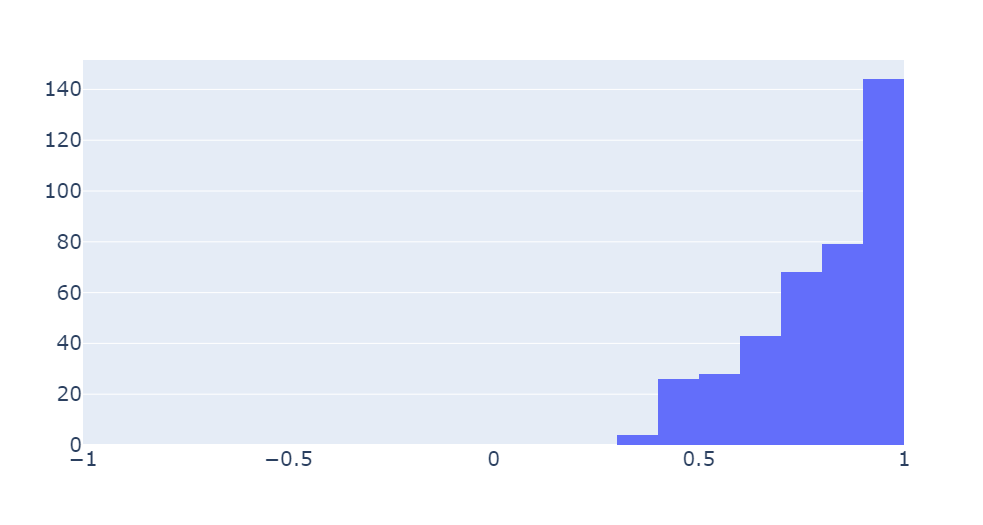}
		\caption{Layer 9}
	\end{subfigure}
		\begin{subfigure}[b]{0.45\textwidth}
		\centering
		\includegraphics[scale=0.22]{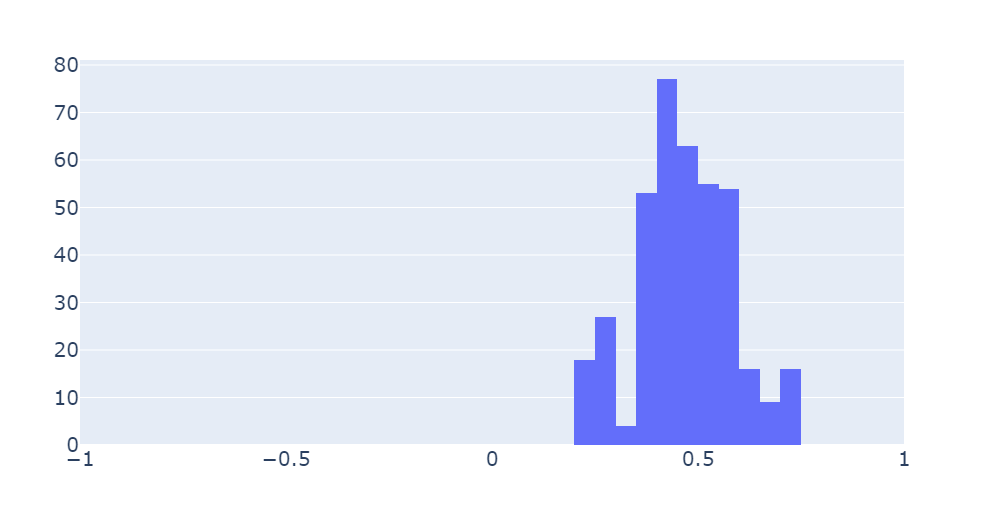}
		\caption{Layer 10}
	\end{subfigure}

	\caption{\revnew{}{Histogram of the gating weights generated by the gating module on videos from SS-v1 dataset.}}
	\label{fig:shift_patterns}
\end{figure*}

\revnew{}{We further analyse the gating weights generated by GSF on BNInception backbone CNN across its various layers. We plot a histogram of the weights averaged across all the videos on Fig.~\ref{fig:shift_patterns}. From the figure, one can see that the gating weights are negative for the bottom layers while it becomes positive at the top layers. The negative weights result in feature differencing while the positive weights result in feature averaging, as explained in Sec.~\ref{sec:gsf_module} of the manuscript. Image difference captures the salient motion occurring between the image pair. Since image features obtained from a CNN are invariant to translation and small changes to appearance, image feature difference allows in encoding of strong temporal information. Once the temporal information is encoded via feature differencing in the bottom layers, the top layers perform feature aggregation via feature averaging. Instead of simply shifting the features as done in TSM, the gating module allows this flexibility of performing feature differencing and averaging to the model, resulting in an improved recognition performance}

\begin{figure*}[h!]
	\centering
	\begin{subfigure}[b]{0.005\textwidth} 
		\raisebox{4in}{\rotatebox[origin=c]{90}{\texttt{folding something}}}
	\end{subfigure} \hskip -7mm
	\begin{subfigure}[b]{0.24\textwidth}
		\centering
		\includegraphics[scale=0.17]{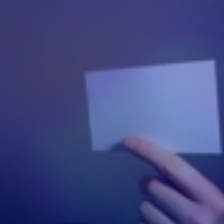}\hskip 1mm\includegraphics[scale=0.17]{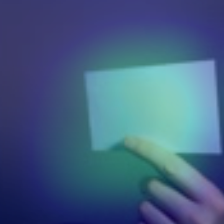} \\[1mm]
		\includegraphics[scale=0.17]{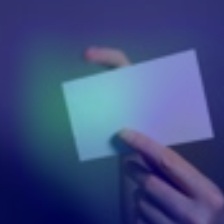}\hskip 1mm\includegraphics[scale=0.17]{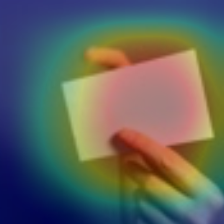} \\[1mm]
		\includegraphics[scale=0.17]{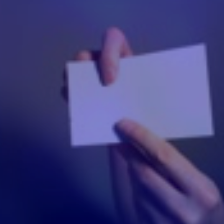}\hskip 1mm\includegraphics[scale=0.17]{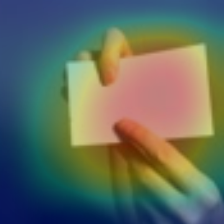} \\[1mm]
		\includegraphics[scale=0.17]{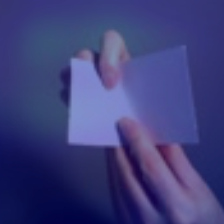}\hskip 1mm\includegraphics[scale=0.17]{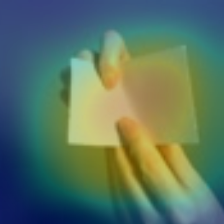} \\[1mm]
		\includegraphics[scale=0.17]{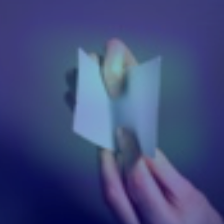}\hskip 1mm\includegraphics[scale=0.17]{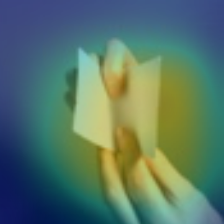} \\[1mm]
		\includegraphics[scale=0.17]{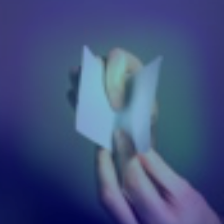}\hskip 1mm\includegraphics[scale=0.17]{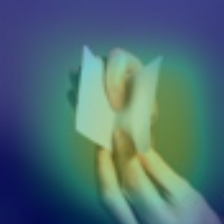} \\[1mm]
		\includegraphics[scale=0.17]{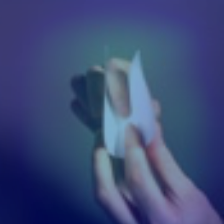}\hskip 1mm\includegraphics[scale=0.17]{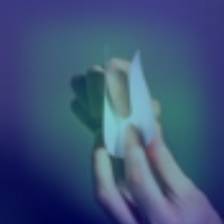} \\[1mm]
		\includegraphics[scale=0.17]{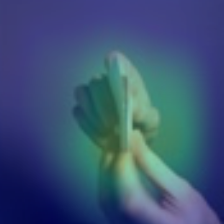}\hskip 1mm\includegraphics[scale=0.17]{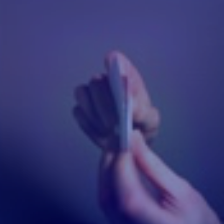} \\[1mm]
		\includegraphics[scale=0.17]{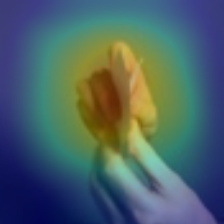}\hskip 1mm\includegraphics[scale=0.17]{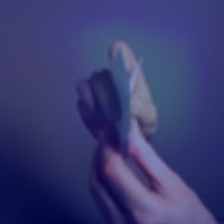} \\[1mm]
		\includegraphics[scale=0.17]{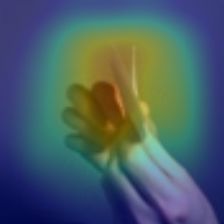}\hskip 1mm\includegraphics[scale=0.17]{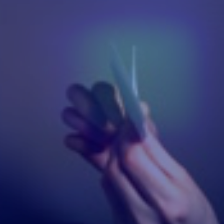} \\[1mm]
		\includegraphics[scale=0.17]{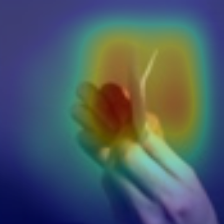}\hskip 1mm\includegraphics[scale=0.17]{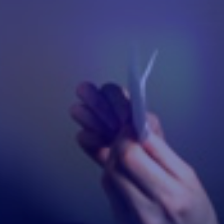} \\[1mm]
		\includegraphics[scale=0.17]{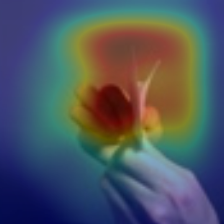}\hskip 1mm\includegraphics[scale=0.17]{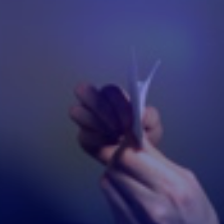} \\[1mm]
		\includegraphics[scale=0.17]{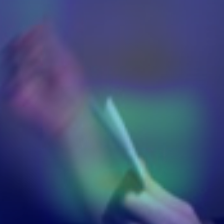}\hskip 1mm\includegraphics[scale=0.17]{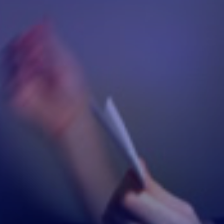} \\[1mm]
		\includegraphics[scale=0.17]{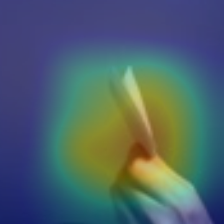}\hskip 1mm\includegraphics[scale=0.17]{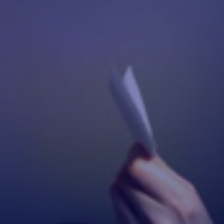} \\[1mm]
		\includegraphics[scale=0.17]{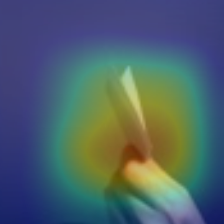}\hskip 1mm\includegraphics[scale=0.17]{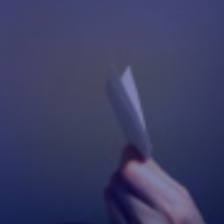} \\[1mm]
		\includegraphics[scale=0.17]{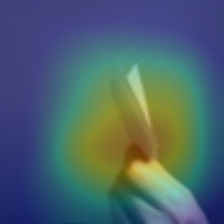}\hskip 1mm\includegraphics[scale=0.17]{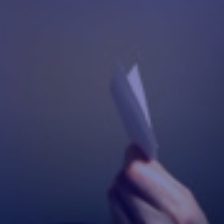} 
		\caption{}
		\label{fig:folding}
	\end{subfigure}
	\begin{subfigure}[b]{0.005\textwidth} 
		\raisebox{4in}{\rotatebox[origin=c]{90}{\texttt{unfolding something}}}
	\end{subfigure} \hskip -7mm
	\begin{subfigure}[b]{0.24\textwidth}
		\centering
		\includegraphics[scale=0.17]{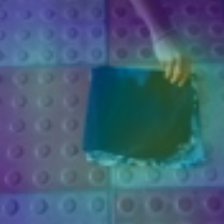}\hskip 1mm\includegraphics[scale=0.17]{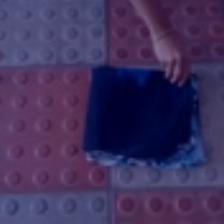} \\[1mm]
		\includegraphics[scale=0.17]{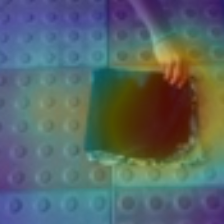}\hskip 1mm\includegraphics[scale=0.17]{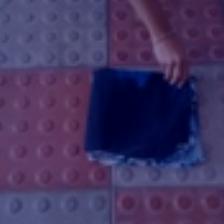} \\[1mm]
		\includegraphics[scale=0.17]{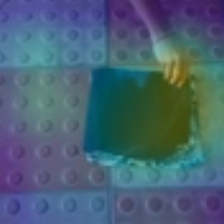}\hskip 1mm\includegraphics[scale=0.17]{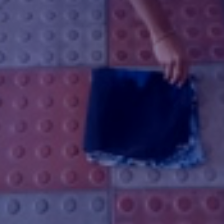} \\[1mm]
		\includegraphics[scale=0.17]{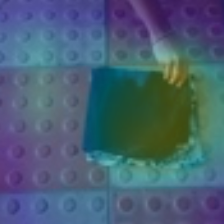}\hskip 1mm\includegraphics[scale=0.17]{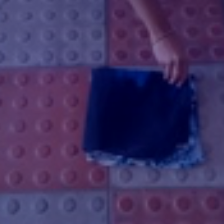} \\[1mm]
		\includegraphics[scale=0.17]{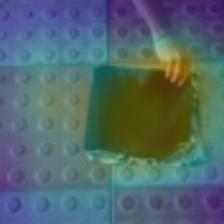}\hskip 1mm\includegraphics[scale=0.17]{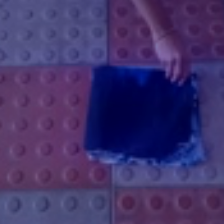} \\[1mm]
		\includegraphics[scale=0.17]{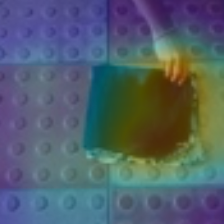}\hskip 1mm\includegraphics[scale=0.17]{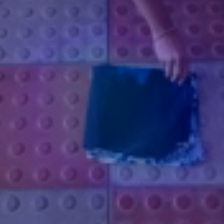} \\[1mm]
		\includegraphics[scale=0.17]{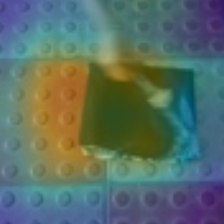}\hskip 1mm\includegraphics[scale=0.17]{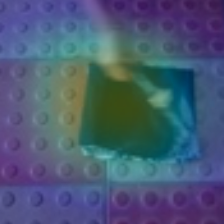} \\[1mm]
		\includegraphics[scale=0.17]{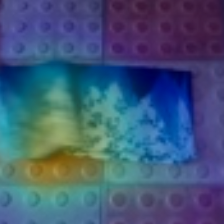}\hskip 1mm\includegraphics[scale=0.17]{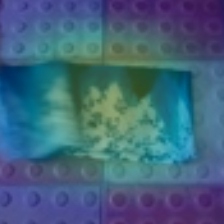} \\[1mm]
		\includegraphics[scale=0.17]{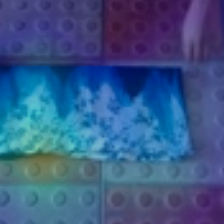}\hskip 1mm\includegraphics[scale=0.17]{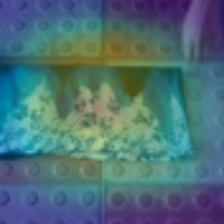} \\[1mm]
		\includegraphics[scale=0.17]{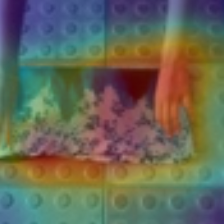}\hskip 1mm\includegraphics[scale=0.17]{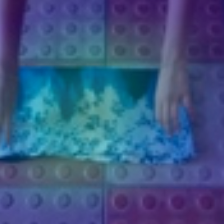} \\[1mm]
		\includegraphics[scale=0.17]{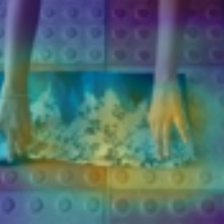}\hskip 1mm\includegraphics[scale=0.17]{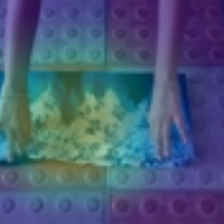} \\[1mm]
		\includegraphics[scale=0.17]{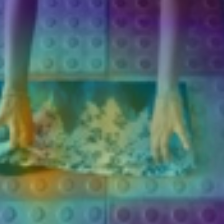}\hskip 1mm\includegraphics[scale=0.17]{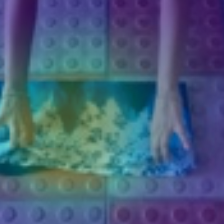} \\[1mm]
		\includegraphics[scale=0.17]{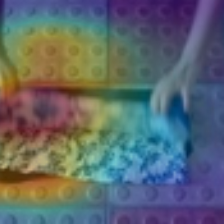}\hskip 1mm\includegraphics[scale=0.17]{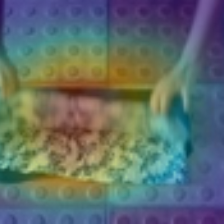} \\[1mm]
		\includegraphics[scale=0.17]{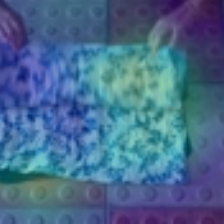}\hskip 1mm\includegraphics[scale=0.17]{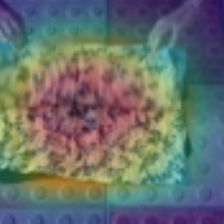} \\[1mm]
		\includegraphics[scale=0.17]{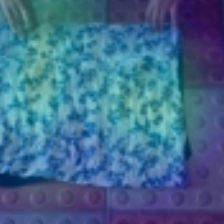}\hskip 1mm\includegraphics[scale=0.17]{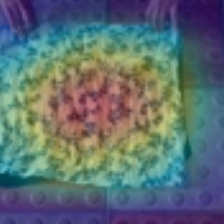} \\[1mm]
		\includegraphics[scale=0.17]{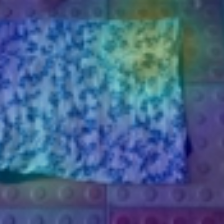}\hskip 1mm\includegraphics[scale=0.17]{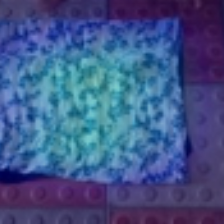} 
		\caption{}
		\label{fig:unfolding}
	\end{subfigure}
	\begin{subfigure}[b]{0.005\textwidth} 
		\raisebox{4in}{\rotatebox[origin=c]{90}{\texttt{plugging something into something but pulling it right out as you remove your hand}}}
	\end{subfigure} \hskip -7mm
	\begin{subfigure}[b]{0.24\textwidth}
		\centering
		\includegraphics[scale=0.17]{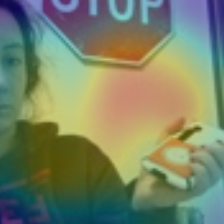}\hskip 1mm\includegraphics[scale=0.17]{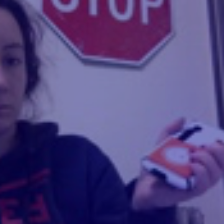} \\[1mm]
		\includegraphics[scale=0.17]{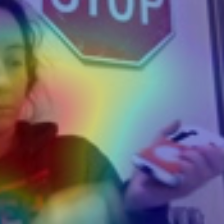}\hskip 1mm\includegraphics[scale=0.17]{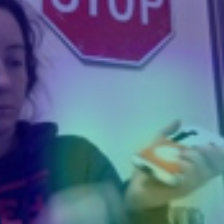} \\[1mm]
		\includegraphics[scale=0.17]{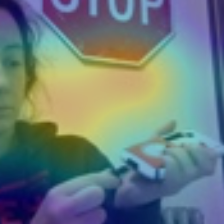}\hskip 1mm\includegraphics[scale=0.17]{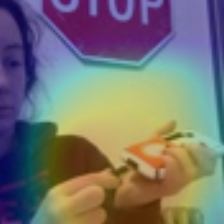} \\[1mm]
		\includegraphics[scale=0.17]{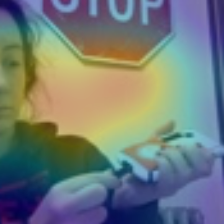}\hskip 1mm\includegraphics[scale=0.17]{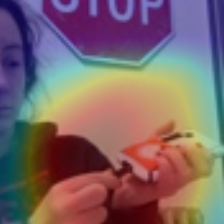} \\[1mm]
		\includegraphics[scale=0.17]{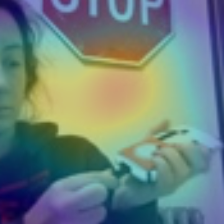}\hskip 1mm\includegraphics[scale=0.17]{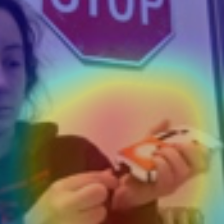} \\[1mm]
		\includegraphics[scale=0.17]{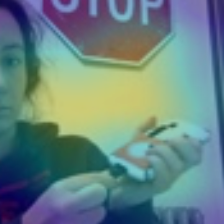}\hskip 1mm\includegraphics[scale=0.17]{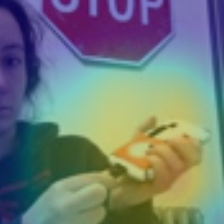} \\[1mm]
		\includegraphics[scale=0.17]{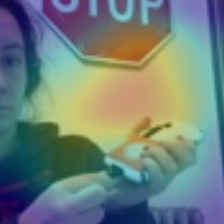}\hskip 1mm\includegraphics[scale=0.17]{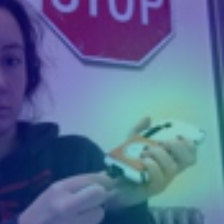} \\[1mm]
		\includegraphics[scale=0.17]{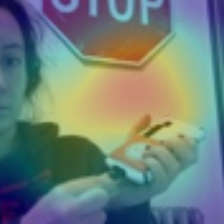}\hskip 1mm\includegraphics[scale=0.17]{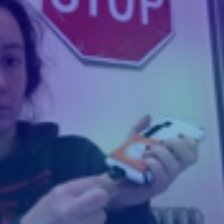} \\[1mm]
		\includegraphics[scale=0.17]{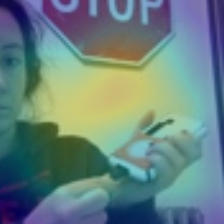}\hskip 1mm\includegraphics[scale=0.17]{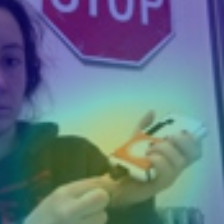} \\[1mm]
		\includegraphics[scale=0.17]{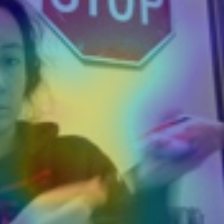}\hskip 1mm\includegraphics[scale=0.17]{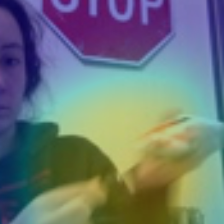} \\[1mm]
		\includegraphics[scale=0.17]{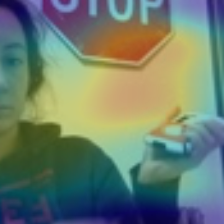}\hskip 1mm\includegraphics[scale=0.17]{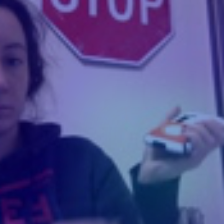} \\[1mm]
		\includegraphics[scale=0.17]{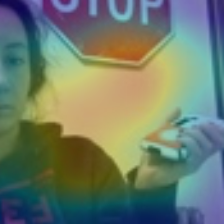}\hskip 1mm\includegraphics[scale=0.17]{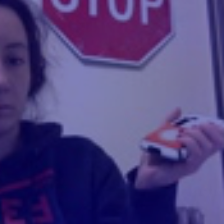} \\[1mm]
		\includegraphics[scale=0.17]{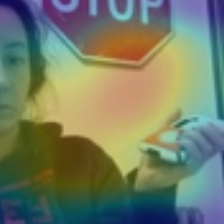}\hskip 1mm\includegraphics[scale=0.17]{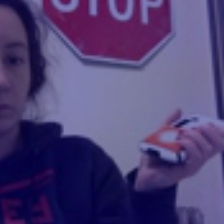} \\[1mm]
		\includegraphics[scale=0.17]{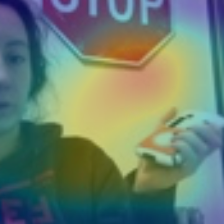}\hskip 1mm\includegraphics[scale=0.17]{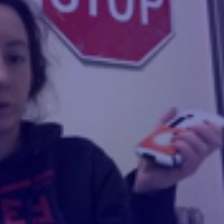} \\[1mm]
		\includegraphics[scale=0.17]{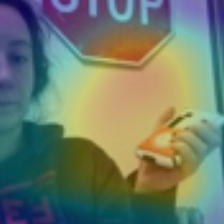}\hskip 1mm\includegraphics[scale=0.17]{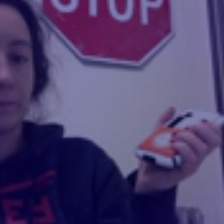} \\[1mm]
		\includegraphics[scale=0.17]{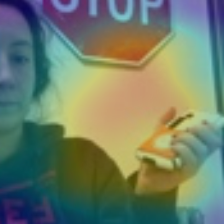}\hskip 1mm\includegraphics[scale=0.17]{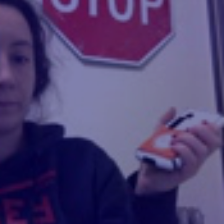} 
		\caption{}
		\label{fig:plugging_and_removing}
	\end{subfigure}
	\begin{subfigure}[b]{0.005\textwidth} 
		\raisebox{4in}{\rotatebox[origin=c]{90}{\texttt{plugging something into something}}}
	\end{subfigure} \hskip -7mm
	\begin{subfigure}[b]{0.24\textwidth}
		\centering
		\includegraphics[scale=0.17]{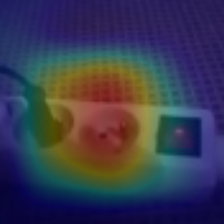}\hskip 1mm\includegraphics[scale=0.17]{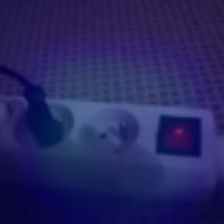} \\[1mm]
		\includegraphics[scale=0.17]{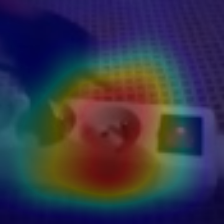}\hskip 1mm\includegraphics[scale=0.17]{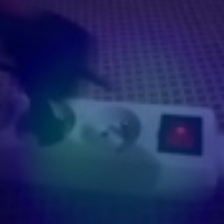} \\[1mm]
		\includegraphics[scale=0.17]{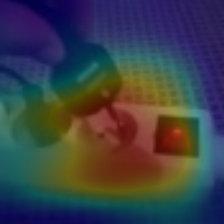}\hskip 1mm\includegraphics[scale=0.17]{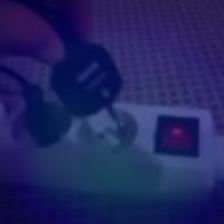} \\[1mm]
		\includegraphics[scale=0.17]{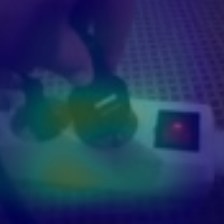}\hskip 1mm\includegraphics[scale=0.17]{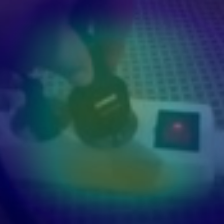} \\[1mm]
		\includegraphics[scale=0.17]{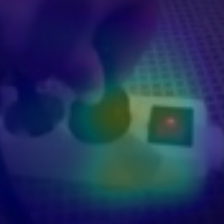}\hskip 1mm\includegraphics[scale=0.17]{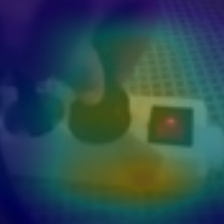} \\[1mm]
		\includegraphics[scale=0.17]{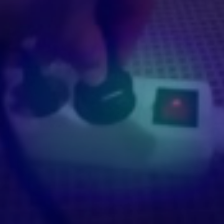}\hskip 1mm\includegraphics[scale=0.17]{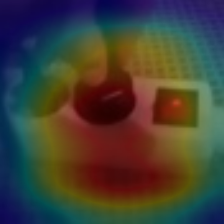} \\[1mm]
		\includegraphics[scale=0.17]{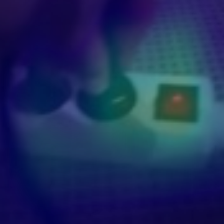}\hskip 1mm\includegraphics[scale=0.17]{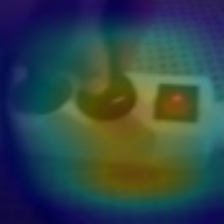} \\[1mm]
		\includegraphics[scale=0.17]{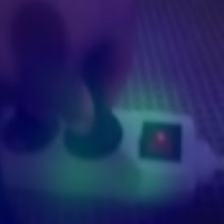}\hskip 1mm\includegraphics[scale=0.17]{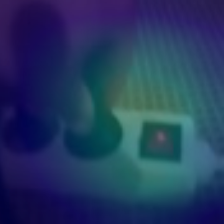} \\[1mm]
		\includegraphics[scale=0.17]{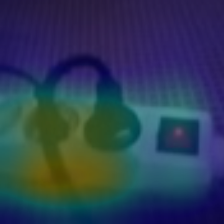}\hskip 1mm\includegraphics[scale=0.17]{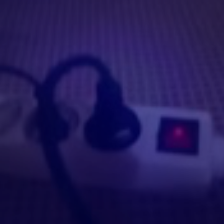} \\[1mm]
		\includegraphics[scale=0.17]{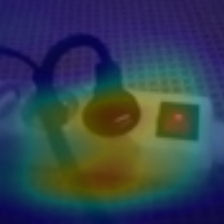}\hskip 1mm\includegraphics[scale=0.17]{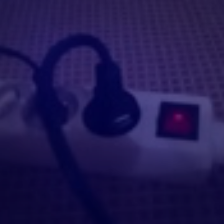} \\[1mm]
		\includegraphics[scale=0.17]{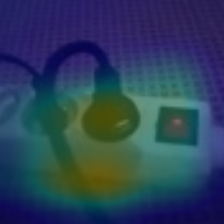}\hskip 1mm\includegraphics[scale=0.17]{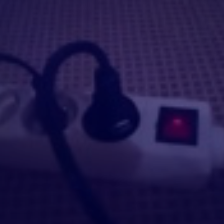} \\[1mm]
		\includegraphics[scale=0.17]{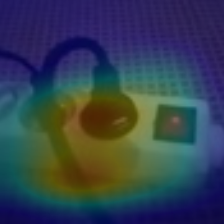}\hskip 1mm\includegraphics[scale=0.17]{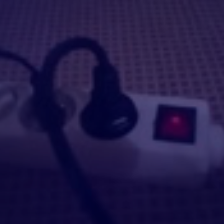} \\[1mm]
		\includegraphics[scale=0.17]{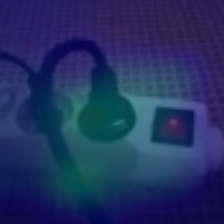}\hskip 1mm\includegraphics[scale=0.17]{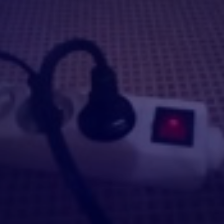} \\[1mm]
		\includegraphics[scale=0.17]{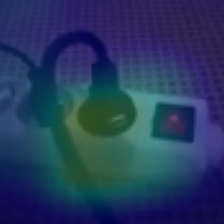}\hskip 1mm\includegraphics[scale=0.17]{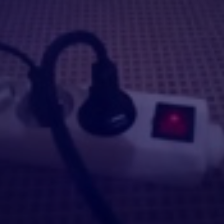} \\[1mm]
		\includegraphics[scale=0.17]{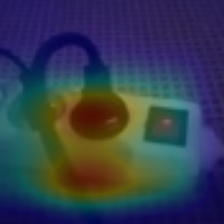}\hskip 1mm\includegraphics[scale=0.17]{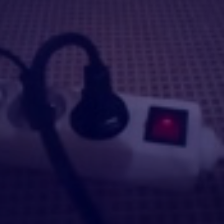} \\[1mm]
		\includegraphics[scale=0.17]{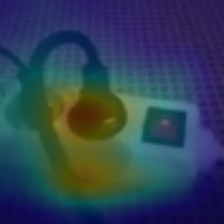}\hskip 1mm\includegraphics[scale=0.17]{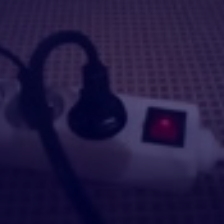} 
		\caption{}
		\label{fig:plugging}
	\end{subfigure}
	\caption{Saliency tubes generated by TSN (left) and GSF (right) on sample videos taken from the validation set of Something Something-V1 dataset. Action labels are shown as text on columns.}
	\label{fig:visualization}
\end{figure*}

The action categories that benefited the most by plugging in \ac{gsf} to the backbone \ac{cnn} are reported in Fig.~\ref{fig:improved_classes}. The Y-axis labels show the true labels (predicted by~\ac{gsf}) and the labels predicted by~\ac{tsn}, separated by ``/". From the plot, it can be seen that the baseline~\ac{tsn} confuses actions that are of similar nature but following a different (reversed) order, such as \texttt{unfolding something} and \texttt{folding something}, \texttt{moving something and something closer to each other} and \texttt{moving something and something away from each other}, \etc With~\ac{gsf} modules in the backbone~\ac{cnn}, the model is capable of encoding the temporal ordering of frames to correctly recognize the action. This can further be seen from the visualization of the saliency tubes~\cite{saliency} for the samples from the most improved classes as shown in Fig.~\ref{fig:visualization}. From the figure, one can see that \ac{gsf} predicts the action category by focusing on the relevant frames where the corresponding action is taking place as opposed to the baseline~\ac{tsn}. In Figs.~\ref{fig:plugging_and_removing} and~\ref{fig:plugging}, even though the baseline model is focusing on the relevant spatio-temporal features, the lack of temporal reasoning renders the model to make wrong predictions. 
\begin{figure}[t]
	\centering
	\includegraphics[width=\columnwidth]{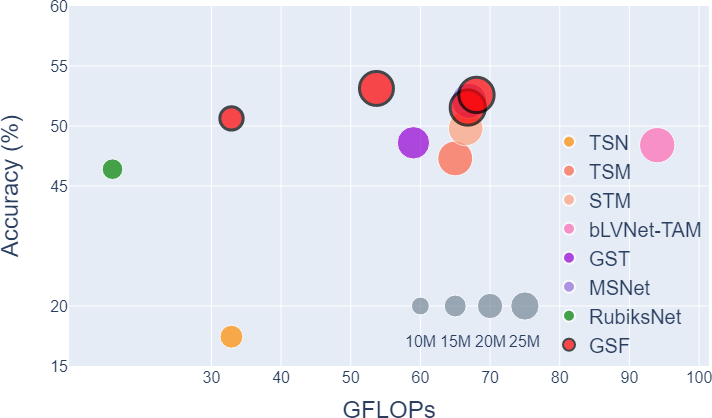}
	\caption{Accuracy (\%) vs complexity (GFLOPs) of \sota approaches on Something-V1 dataset.}
	\label{fig:bubble_plot}
\end{figure}

We analyse the computational complexity and memory requirement for \acp{gsf} and \sota models in Fig.~\ref{fig:bubble_plot}. X-axis shows the computational complexity in GFLOPs and Y-axis shows the recognition accuracy. The size of the marker indicate the number of parameters present in each model. It can be seen from the plot that \ac{gsf} outperforms existing approaches in terms of recognition accuracy while maintaining a reasonable compute and memory requirements.\revnew{}{Tab.~\ref{tab:latency} compares the inference latency of GSF against TSM and GST. We report the time taken for a forward pass with a single input video consisting of 8 frames. We measure the latency on a single NVIDIA Titan V GPU. Compared to the other two architectures, GSF is slightly slow with a latency of 12.4ms compared to 7.59ms of TSM \revnewv{and}{,} 8.28ms of GST and \revnewv{}{10.46ms of RubiksNet}. However, this latency of 12.4ms is equivalent to $\approx$80 videos per second, which is a reasonable speed for real time applications. On the other hand, the improvement in accuracy obtained by GSF on SS-V1 is large compared to the other \revnewv{two}{three} architectures (+4.57\% over TSM \revnewv{and}{,} +3.17\% over GST \revnewv{}{and +3.77\%} over RubiksNet).} \revnewv{}{Note that all models except RubiksNet are run with half precision. We report additional runtime analyses in the supplementary document.}

\begin{table}[t]
    \centering
    \begin{tabular}{|c|c|c|c|}
    \hline
         \textbf{Model} & \textbf{Latency (ms)} &  \textbf{Accuracy on SS-V1 (\%)}   \\ \hline \hline
         TSM & 7.59 &  45.6\\
         GST & 8.28 &  47.0\\ 
         \revnewv{}{RubiksNet}$^*$ & \revnewv{}{10.46} & \revnewv{}{46.4} \\ \hline
         GSF & 12.40 &  50.17\\ \hline
    \end{tabular}
    \caption{\revnew{}{Comparison of latency incurred by GSF with respect to other efficient approaches. We report the time taken for one forward pass of a video clip consisting of 8 frames. }\revnewvt{}{$^*$: All models except RubiksNet are run with half precision.}}
    \label{tab:latency}
\end{table}
		
		\section{Conclusion}
		\label{sec:conclusion}
		
		\newtext{We presented \acf{gsf}, a novel spatio-temporal feature extraction  module capable of converting a 2D~\ac{cnn} into a high performing spatio-temporal feature extractor with negligible computational overhead. \ac{gsf} relies on spatial gating and channel fusion to exchange information between neighbouring frames in a data dependent manner. We performed an extensive analysis to study the effectiveness of~\ac{gsf} using two \ac{cnn} model families on five standard action recognition benchmarks. Compared to other approaches, \ac{gsf} achieves \sota or competitive performance on these datasets with far less model complexity, indicating its effectiveness as an efficient spatio-temporal feature extractor.}

		\section*{Acknowledgements}
		This work has been partially supported by the Spanish projects PID2019-105093GB-I00, TED2021-131317B-I00 and PDC2022-133305-I00. We gratefully acknowledge the support from Amazon AWS Machine Learning Research Awards (MLRA), ICREA under the ICREA Academia programme, start-up grant IN2814 of Free University of Bozen-Bolzano and NVIDIA AI Technology Centre (NVAITC), EMEA. We acknowledge the CINECA award under the ISCRA initiative, for the availability of high performance computing resources and support.

		\bibliographystyle{ieee}
		\bibliography{gsf_bib}
			
		
	\end{document}